\documentclass[10pt,journal,cspaper,compsoc]{IEEEtran}
\ifCLASSOPTIONcompsoc
\else
\fi

\ifCLASSINFOpdf
\else
\fi

\usepackage[T1]{fontenc}
\usepackage[latin1]{inputenc}
\usepackage{verbatim}
\usepackage{float}
\usepackage{amsmath}
\usepackage{graphicx}
\usepackage{amssymb}
\usepackage{cite}
\makeatletter

\floatstyle{ruled}
\newfloat{algorithm}{tbp}{loa}
\floatname{algorithm}{Algorithm}

\usepackage{float}
\usepackage{subfloat}
\usepackage{amsfonts}
\usepackage{algorithm}
\usepackage{algorithmic}

\makeatletter

\floatstyle{ruled}
\newfloat{algorithm}{tbp}{loa}
\floatname{algorithm}{Algorithm}
\usepackage{setspace}
\usepackage{accents}
\usepackage{hyperref}
\usepackage[symbol]{footmisc}

\floatstyle{ruled}
\newfloat{algorithm}{tbp}{loa}
\floatname{algorithm}{Algorithm}
\usepackage{subfigure}

\providecommand{\U}[1]{\protect\rule{.1in}{.1in}}

\makeatother

\begin{document}
%
\title{The Sum-over-Forests density index: \\
identifying dense regions in a graph}


%
%
%
%

\author{Mathieu Senelle, Silvia~Garcia-Diez, Amin~Mantrach, Masashi~Shimbo,  Marco~Saerens \& François~Fouss%
\IEEEcompsocitemizethanks{\IEEEcompsocthanksitem A. Mantrach is with IRIDIA Laboratory, Universit\'{e} Libre de Bruxelles, Brussels, Belgium. Email: amantrac@ulb.ac.be.
\IEEEcompsocthanksitem M. Senelle, S. Garcia-Diez, M. Saerens and F. Fouss are with the Louvain School of Management \& Machine Learning Group, Universit\'{e} catholique de Louvain (UCL), Belgium. Email:
m.senelle@uclouvain-mons.be,
\{silvia.garciadiez, marco.saerens, françois.fouss\}@uclouvain.be.
\IEEEcompsocthanksitem M. Shimbo is with the Graduate School of Information Science, Nara Institute of Science and Technology, Japan. Email: shimbo@is.naist.jp.}%

}
\IEEEcompsoctitleabstractindextext{%
\begin{abstract}
This work introduces a novel nonparametric density index defined on graphs, the Sum-over-Forests (SoF) density index. It is based on a clear and intuitive idea: high-density regions in a graph are characterized by the fact that they contain a large amount of low-cost trees with high outdegrees while low-density regions contain few ones. Therefore, inspired by \cite{Mantrach2010}, a Boltzmann probability distribution on the countable set of forests in the graph is defined so that large (high-cost) forests occur with a low probability while short (low-cost) forests occur with a high probability. Then, the SoF density index of a node is defined as the expected outdegree of this node in a non-trivial tree of the forest, thus providing a measure of density around that node. Following the matrix-forest theorem \cite{Chebotarev-1997,Chebotarev-2002b} and a statistical physics framework, it is shown that the SoF density index can be easily computed in closed form through a simple matrix inversion.  Experiments on artificial and real data sets show that the proposed index performs well on finding dense regions, for graphs of various origins. 
\end{abstract}

}

\maketitle

\IEEEdisplaynotcompsoctitleabstractindextext

%
\IEEEpeerreviewmaketitle

\section{Introduction}
%
%

%
%
%
%
\subsection{General introduction}
\IEEEPARstart{D}{ensity} is an important concept in graph analysis and has been proven to be of particular interest in various areas such as, for example, social networks, biology and World-Wide-Web \cite{Luce1949, Li2007, Flake2000}.

The task of identifying dense regions on a graph can be based on various concepts (degree of a node, cliques, cores, etc.) leading to various approaches (see Section \ref{sec:related work}). The key concept on which our approach is based is forest enumeration and, in particular, the matrix-forest theorem \cite{Chebotarev-1997,Chebotarev-2002b}, an extension of the well-known matrix-tree theorem (see, e.g., \cite{Tutte-2001}). More precisely, the method developed in this paper, inspired by \cite{Akamatsu-1996,Saerens-2008,Yen-08K,Mantrach2010} (based on paths instead of forests), relies on the enumeration of all the possible forests in the graph, therefore leading to the definition of a new density index which will be called the \textbf{Sum-over-Forests} (SoF) \textbf{density index}. This measure has a clear and intuitive interpretation: when enumerating all the possible forests in the graph, a node will be considered as having a high density index if it is part of a tree of many -- preferably low-cost -- forests, and has a high outdegree within this forest. Indeed, if a region has a high density, it will contain a large number of trees -- and therefore forests -- so that the nodes belonging to that region will be part of many forests and have a high outdegree. Those nodes will thus obtain a high SoF density index.

In order to compute this index, we first define a Boltzmann probability distribution on the countable set of forests in the graph by adopting a statistical physics framework. This distribution has the desired property that high-cost forests occur with a low probability while low-cost forests occur with a high probability. As in statistical physics, it depends on a parameter, $\theta=1/T$, controlling the temperature $T$ -- and thus the entropy -- of the system. When $T$ is low, only low-cost forests are taken into account (high-cost forest having a negligible contribution) while for high values of $T$, high-cost forests are as important as the low-cost ones (uniform distribution).

In a second step, the SoF density index of a node is defined according to this probability distribution. Roughly speaking, it corresponds to the expectation of the outdegree of this node, averaged over all the forests (the expectation is taken on all the possible forests). Technically speaking, the SoF density index is obtained by taking the first-order derivative of the partition function associated to the system. It is shown that it can be computed in closed form by inverting a $n\times n$ matrix depending on the immediate costs assigned to the arcs.

\subsection{Related work}\label{sec:related work}

This section provides a short survey of the related work aiming at finding dense regions on graphs. 

A well-known approach for finding high density regions on graphs relies on identifying dense, highly connected subgraphs like cliques, plexes, cores, etc. (see, e.g., \cite{Brandes2005}). Cliques are completely connected subgraphs of the original graph \cite{Luce1949}. Unfortunately, finding all the cliques, or the maximal clique in a graph is NP-complete. As the notion of clique is very restrictive (if an arc is missing, then the subgraph is no more considered as a clique), other ideas relaxing this notion appeared, such as plexes \cite{Seidman1978}. A k-plex is a subgraph
containing $n$ nodes where each node is connected to at least $n-k$ other nodes. Finding k-plex is alas as hard as finding cliques \cite{Brandes2005}. Cores are similar
to plexes, but instead of specifying how many links are missing to produce a clique, nodes inside k-core only have to present a degree superior to $k$ \cite{Newman-2010}.
All nodes of the core are then connected to at least $k$ other members of the core. Contrarily to cliques and plexes, cores can be computed in 
polynomial time, and there even exists linear-time algorithm computing the core structure of a network \cite{Batagelj2010}. A generalization of the notion of core, called the generalized k-core, is based on other vertex properties than the degree (in/out degree, clustering coefficient,...) and can also be found in \cite{Batagelj2010}. Our SoF density index could be used in conjunction with a generalized k-core.

Density-based clustering methods use a measure of density on graphs as an intermediary step for computing clusters. DBSCAN \cite{Ester1996}, a widely used clustering algorithm, computes the local density around a node as the number of neighbours in a sphere of a certain radius around that node. Mode-seeking methods, like Mean Shift \cite{Koontz1976}, compute the modes of a probability density function to find high density areas. These methods were originally intended to be used in the feature space of the data, but adaptations to graph data were recently proposed \cite{Falkowski2007, Jouili2010, Liu2010, Cho2012}. 

Another approach for finding dense zones is to compute a density index (or score) on the nodes of a graph. One of the most intuitive density index is the degree of a node (on undirected graphs, in/out degree on directed graphs) defined as the number of links a node has. Indeed, the larger the number of neighbours of a node, the higher the density around it. This measure is then purely local, taking only into account the direct neighbours. The strength of a node is an extension of the degree to weighted graphs, computing the sum of the weights borne by the arcs of the neighbouring nodes. When those weights are all equal to one, the strength reduces to the degree. The clustering coefficient \cite{Watts-1998} of a node $i$ is also a notion related to the degree. It counts the number of connected neighbours of $i$, divided by the total number of possible connections between those neighbours. This measure was extended to weighted graphs in \cite{Barrat-2004}. 

Similarly, the Sum-over-Forests (SoF) density index developped in this paper computes a density score on nodes by enumerating forests on a graph using the matrix forest theorem \cite{Chebotarev-1997}. This method is based on a sum-over-forests statistical physics framework.

\subsection{Contributions and organization of the paper}

This work has three main contributions:

\begin{itemize}
\item It defines a new density index on nodes of a directed graph.  
\item It shows how this density index can be computed efficiently through a statistical physics framework from the immediate costs associated to each arc by inverting a $n\times n$ matrix. 
\item It shows through experiments on artificial and real data sets that the SoF index is an accurate tool for identifying dense regions on graphs. 
\end{itemize}
Section 2 introduces the necessary background and notation. In Section 3, the probability distribution on the set of forests -- a Boltzmann
distribution -- is defined. Section 4 introduces our index and shows how it can be derived analytically
from the partition function. Section 5 explains how the partition function can be computed exactly from the immediate costs while Section 6 derives
the formulas for computing the density index. Section 7 applies the index to the identification of dense areas
on graphs from various origin. Concluding remarks and possible extensions are discussed in Section 8.

\section{Background and notation}

Consider a weighted directed graph or network without self-loops, $G$, not necessarily strongly connected, with a set of $n$ nodes $V$ (or vertices) and a set of arcs $E$
(or edges). To each arc linking node $k$ and $k'$, we associate a positive number $c_{kk'}>0$ representing the \textbf{immediate cost} of following
this arc. The \textbf{cost matrix} $\mathbf{C}$ is the matrix containing the immediate costs $c_{kk'}$ as elements.
If, instead of $\mathbf{C}$, we are given an adjacency matrix with elements $a_{kk'} \ge 0$ indicating the affinity between node $k$ and
node $k'$, the corresponding costs could be computed from $c_{kk'} = 1/a_{kk'}$. Notice, however, that other relations
-- other than the reciprocal relation -- between affinity and cost could be considered as well. The adjacency matrix containing the elements $a_{kk'}$ is denoted by $\mathbf{A}$, while the Laplacian matrix of a graph having adjacency matrix $\mathbf{A}$ is $\mathbf{L}(\mathbf{A}) = \mathbf{D}-\mathbf{A}$, where $\mathbf{D} = \mathbf{Diag}(\mathbf{A}^{\text{T}}\mathbf{e})$ is a diagonal matrix containing the column sums of $\mathbf{A}$. Here, $\mathbf{e}$ is a column vector full of 1's. Moreover, if the graph is undirected, it is assumed that, for each arc, there exists directed links in the two directions $k \rightarrow k'$ and $k' \rightarrow k$.

The objective of the next sections is to define the probability distribution on the set of forests as well as the density index. 
Before diving into the details, let us briefly describe the main ideas behind the model. In a first step, the set of forests in the graph
is enumerated through the matrix-forest theorem and a probability distribution is assigned to each individual forest: the larger the forest,
the smaller the weight of its contribution, given that isolated nodes do not contribute. 
This probability distribution depends on a parameter, $\theta=1/T$, controlling the smoothing level
carried out in the graph: when $\theta$ is large, only the lowest-cost forests are considered while when
$\theta$ is small, higher-cost forests are also taken into account. In a second step, the expected outdegree each node takes in a forest is 
computed through a sum-over-forests statistical physics formalism, providing a measure of density on the set of nodes.

\section{A Boltzmann distribution on the set of forests}

The present section describes how the probability distribution on the set of forests is assigned. To this end, let us define the set of rooted
forests $\varphi$ that can be defined in the graph $G$ as $\mathcal{F} = \{\varphi_1, \varphi_2, \dots\}$. Intuitively, a rooted forest is an acyclic subgraph of $G$ that has the same nodes as $G$ and one marked node (a root) in each component (see \cite{Chebotarev-1997,Chebotarev-2002b} for details). In the directed case, diverging forests are considered, that is, forests containing diverging rooted trees (i.e., trees that contain only directed paths from the root to all the other nodes). Now, as we are dealing with directed graphs, diverging rooted trees and forests will simply be referred to as trees and forests. The \emph{total cost} of such a forest $\varphi$ is defined as the \emph{sum} of the individual costs of the arcs
belonging to $\varphi$, $C(\varphi)$. On the other hand, the \emph{total weight} of such a forest $\varphi$ is defined as the \emph{product} of the individual weights (the elements of the adjacency matrix) of the arcs belonging to $\varphi$. A forest with no arc (containing only individual nodes without any connection) has a 0 total cost and a total weight of 1.

A \textbf{Boltzmann probability distribution} is defined on the set $\mathcal{F}$:
\begin{align}
\text{P}(\varphi) & =\frac{\exp\left[-\theta C(\varphi)\right]}{{\displaystyle \sum_{\varphi\in\mathcal{F}}} \exp\left[-\theta C(\varphi)\right]}\label{Eq_Probability_distribution01}
\end{align}
where $\theta$ is the inverse temperature.
Thus, as expected, low-cost forests $\varphi$ (having small $C(\varphi)$) are favored in that they have a large
probability of being chosen. Indeed, from Equation (\ref{Eq_Probability_distribution01}), we clearly observe that when $\theta \rightarrow 0^{+}$, the forest probabilities tend
to a uniform probability. On the other hand, when $\theta$ is large, the probability distribution defined by
Equation (\ref{Eq_Probability_distribution01}) is biased towards low-cost forests (the most likely forests are the lowest-cost ones).
Notice that in Equation (\ref{Eq_Probability_distribution01}) isolated nodes (with no ingoing or outgoing link) do not contribute to the probability.
In the sequel, it will be assumed that the user provides the
value of the parameter $\theta$.

\begin{figure}
\begin{centering}
\includegraphics[scale=0.5]{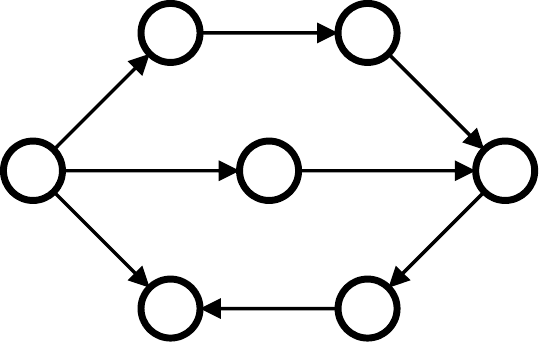}
\par\end{centering}

\caption{\textbf{\scriptsize A directed graph $G$ in which arc costs are uniformly 1.}}
\label{fig:simplegraph}
\end{figure}

\begin{figure}
  \centering
  \subfigure[High-cost forest $\varphi_1$.]{\label{fig:arc-a}\includegraphics[scale=0.5]{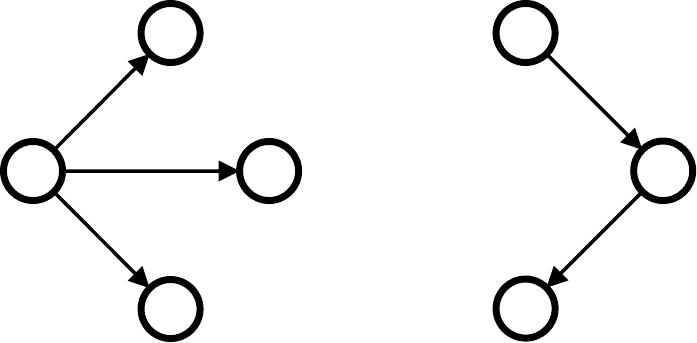}}\qquad                
  \subfigure[Low-cost forest $\varphi_2$.]{\label{fig:contour-b}\raisebox{5mm}{\includegraphics[scale=0.5]{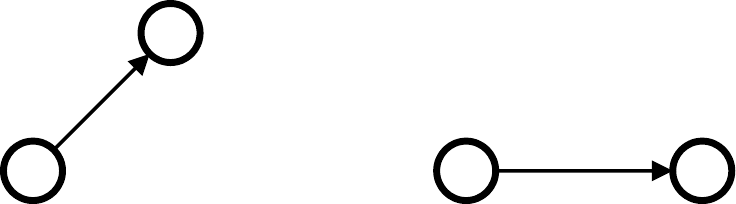}}}
  \caption{\textbf{\scriptsize Examples of forests on graph $G$ containing two trees. Isolated nodes are not displayed since they do not contribute to the density index.}}
  \label{fig:Forests}
\end{figure}

For illustration, the simple graph $G$ shown in Figure \ref{fig:simplegraph} is analysed. Figure \ref{fig:Forests} represents examples of respectively a high-cost forest $\varphi_1$ and a low-cost forest $\varphi_2$ on $G$. The cost associated to $\varphi_1$ is 5, as this forest contains five arcs with a cost equal to 1. Similarly, the cost of $\varphi_2$ is 2. The numerator of the Equation (\ref{Eq_Probability_distribution01}) for $\varphi_1$ becomes $\exp\left[-\theta 5\right]$, the numerator for $\varphi_2$ $\exp\left[-\theta 2\right]$, while the denominator is the same for both forests. For small values of $\theta$, those numerators tend to 1 and the probabilities to the uniform distribution. For high values of $\theta$, the probability of the lower-cost forest $\varphi_2$ is higher than the probability of the higher-cost forest $\varphi_1$.

\section{The SoF density index}

By following arguments inspired from \cite{Saerens-2008}, it is now shown that 
the density index can be computed from a quantity appearing in the denominator of 
Equation (\ref{Eq_Probability_distribution01}), defined as \begin{equation}
\mathcal{Z}=\sum_{\varphi\in\mathcal{F}}\exp\left[-\theta C(\varphi)\right],\label{Eq_Partition01}
\end{equation}
and which corresponds to the \textbf{partition function} in statistical
physics (see \cite{Jaynes-1957} or any textbook in statistical physics;
for instance \cite{Reichl-1998,Schrodinger-1952}). For this purpose, let us further define the \textbf{free energy}
$F$ in the usual way \cite{Reichl-1998,Schrodinger-1952} as \begin{eqnarray} F=-\frac{1}{\theta}\log(\mathcal{Z})=-T\log(\mathcal{Z}) \label{eq:free_energy}\end{eqnarray}
where $T=1/\theta$ is the \textbf{temperature} of the system.
The \textbf{expected number of times} a link $k\rightarrow k'$ is present in a forest can easily be computed through
\begin{align}
\overline{\eta}(k,k') & =\frac{\partial F}{\partial c_{kk'}}=-\frac{1}{\theta}\frac{\partial(\log\mathcal{Z})}{\partial c_{kk'}} \label{eq:firstorderderivateZ} \\
 & =\sum_{\varphi\in\mathcal{F}}\dfrac{\exp\left[-\theta C(\varphi)\right]}{\mathcal{Z}}\delta(\varphi;k,k') \\
 & =\sum_{\varphi\in\mathcal{F}}\text{P}(\varphi)\,\delta(\varphi;k,k')
 \label{eq:firstorderderivateZ02}
 \end{align}
where $\delta(\varphi;k,k')$ is a Kronecker delta indicating if the link
$k\rightarrow k'$ is present in forest $\varphi$, and thus if
the link is part of forest $\varphi$.
The \textbf{expected outdegree} of node
$k$ on a  forest, which defines the \textbf{SoF density index}, is
\begin{equation}
\text{dens}(k)
 =\sum_{\varphi\in\mathcal{F}}\text{P}(\varphi) \left( \sum_{k'=1}^{n} \delta(\varphi;k,k') \right)
 =\sum_{k'=1}^{n}\overline{\eta}(k,k')\label{Eq_Betweenness01}
\end{equation}
and corresponds to the sum of the contributions of the arcs issued from node $k$. 

In the next section, we show that the partition function can easily be computed from the cost matrix.

\section{Computation of the partition function \texorpdfstring{$\mathcal{Z}$}{Z}}

By using the matrix-forest theorem \cite{Chebotarev-1997,Chebotarev-2002b}, let us now show how the partition function
$\mathcal{Z}$ (Equation (\ref{Eq_Partition01})) can be computed exactly from the immediate costs.
Indeed, let us assume a graph $G$ characterized by an adjacency matrix $\mathbf{A}$ containing the weights on the arcs.
From the matrix-forest theorem (see \cite{Chebotarev-1997}, lemma 2, or \cite{Chebotarev-2002b} for details), $\det(\mathbf{I} + \mathbf{L}(\mathbf{A}))$
is the \emph{sum} of the total weights of all the rooted (diverging in the directed case) forests $\varphi\in\mathcal{F}$ that can be extracted from the graph. The total weight of a particular rooted forest $\varphi$ is the \emph{product} of the weights of the individual arcs defining it.

Let us now apply this concept to a new matrix $\mathbf{W}$ defined from the cost matrix, $\mathbf{C}$,
\begin{equation}
\mathbf{W} = \exp\left[-\theta\mathbf{C}\right]\text{,} \label{Eq_W_matrix01}
\end{equation}
where the logarithm/exponential functions are taken elementwise. Thus, the elements of
matrix $\mathbf{W}$ are $\exp\left[-\theta c_{kk'}\right]$. Now, if we set as adjacency
matrix $\mathbf{A} = \mathbf{W}$, the total weight of a rooted forest $\varphi$ is the product of the individual weights defining it, i.e,
$\prod_{k,k':k \rightarrow k' \in \varphi} a_{kk'} = \prod_{k,k':k \rightarrow k' \in \varphi} \exp\left[-\theta c_{kk'}\right] = \exp [-\theta \sum_{k,k':k \rightarrow k' \in \varphi} c_{kk'} ] = \exp\left[-\theta C(\varphi)\right]$.
We can immediately deduce from the matrix-forest theorem that $\det(\mathbf{I} + \mathbf{L}(\mathbf{W}))$,
where $\mathbf{L}(\mathbf{W}) = \mathbf{Diag}(\mathbf{W}^{\text{T}}\mathbf{e}) - \mathbf{W}$, is equal to
$\sum_{\varphi\in\mathcal{F}}\exp\left[-\theta C(\varphi)\right] = \mathcal{Z}$.
Therefore,
\begin{equation}
\mathcal{Z} = \det(\mathbf{I} + \mathbf{L}(\mathbf{W})),\text{ with }\mathbf{W} = \exp\left[-\theta\mathbf{C}\right]
\label{Eq_W_computation01}
\end{equation} 

This result is used in next section in order to derive the SoF density index.

\section{Computation of the SoF density index}

Now that we have seen how to compute the partition function $\mathcal{Z}$,
we turn to the computation of the density index that can be deduced from $\mathcal{Z}$ thanks to Equations
(\ref{eq:firstorderderivateZ}) and (\ref{Eq_Betweenness01}).

We thus have to compute the derivatives of $\mathcal{Z}$ (Equation (\ref{Eq_W_computation01})) in terms of $c_{kk'}$ (see Equation (\ref{eq:firstorderderivateZ})) in order to obtain the different quantities of interest. Now, it is well-known (see, e.g., \cite{Harville-97,Schott-2005}) that $\partial \log(\det(\mathbf{X}))/\partial t = \text{trace}(\mathbf{X}^{-1}\frac{\partial \mathbf{X}}{\partial t})$. Thus, for the expected number of times the link $k\rightarrow k'$ appears in a forest, we obtain
\begin{align}
\overline{\eta}(k,k') &= \frac{\partial F}{\partial c_{kk'}} = -\frac{1}{\theta} \frac{\partial \log(\det(\mathbf{I} + \mathbf{L(W)}))}{\partial c_{kk'}} \nonumber \\
&= -\frac{1}{\theta} \text{trace}(\mathbf{Z} \frac{\partial (\mathbf{I} + \mathbf{L(W)})}{\partial c_{kk'}})\label{Eq_Computation_transitions01} \nonumber \\
&= -\frac{1}{\theta} \text{trace}(\mathbf{Z} \frac{\partial \mathbf{L}(\mathbf{W})}{\partial c_{kk'}}) \nonumber \\
&= -\frac{1}{\theta} \text{trace}(\mathbf{Z} \frac{\partial (\mathbf{D} - \mathbf{W})}{\partial c_{kk'}})
\end{align}
where the matrix $\mathbf{Z}$ is defined as
\begin{equation}
\mathbf{Z} = (\mathbf{I} + \mathbf{L}(\mathbf{W}))^{-1} = (\mathbf{I} + (\mathbf{Diag}(\mathbf{W}^\text{T}\mathbf{e}) - \mathbf{W}))^{-1}
\label{Eq:FundamentalMatrix01}
\end{equation}

Now, we easily find that $\partial \mathbf{W}/\partial c_{kk'} = -\theta w_{kk'} \mathbf{e}_k \mathbf{e}_{k'}^{\text{T}}$ and $\partial \mathbf{D}/\partial c_{kk'} = -\theta w_{kk'} \mathbf{e}_{k'} \mathbf{e}_{k'}^{\text{T}}$ so that
\begin{equation}
\frac{\partial \mathbf{L(\mathbf{W})}}{\partial c_{kk'}} = \frac{\partial (\mathbf{D} - \mathbf{W})}{\partial c_{kk'}} = -\theta w_{kk'} (\mathbf{e}_{k'} \mathbf{e}_{k'}^{\text{T}} - \mathbf{e}_k \mathbf{e}_{k'}^{\text{T}}) \label{Eq_Derivative_laplacian01},
\end{equation}
where $\mathbf{e}_{k}$ is a basis column vector with zeroes everywhere except in position $k$ where there is a $1$.

Thus, by defining $\mathbf{z}_k = \mathbf{col}_{k} (\mathbf{Z})$ as column $k$ of matrix $\mathbf{Z}$,
\begin{align}
\overline{\eta}(k,k') &= \text{trace}(w_{kk'} \mathbf{Z} (\mathbf{e}_{k'} \mathbf{e}_{k'}^{\text{T}} - \mathbf{e}_{k} \mathbf{e}_{k'}^{\text{T}})) \nonumber \\
&= w_{kk'} \text{trace}(\mathbf{z}_{k'} \mathbf{e}_{k'}^{\text{T}}) - \text{trace}(\mathbf{z}_k \mathbf{e}_{k'}^{\text{T}}) \nonumber \\
&= w_{kk'} z_{k'k'} - w_{kk'}z_{k'k} \label{Eq_Computation_between01}
\end{align}

Therefore, the expected outdegree of node $k$ 
-- the \textbf{SoF density index} of node $k$ -- is
\begin{equation}
\text{dens}(k)=\sum_{k'=1}^{n}\overline{\eta}(k,k')=\sum_{k'=1}^{n}(w_{kk'} z_{k'k'} - w_{kk'} z_{k'k}) \label{Eq_Computation_passage01}
\end{equation}
where we used Equations (\ref{Eq_Betweenness01}) and (\ref{Eq_Computation_between01}). The $n \times 1$
column vector containing the elements $\text{dens}(k)$ will be called
$\mathbf{d}$, with
\begin{equation}
\mathbf{d} = \mathbf{W} \, \textbf{\text{diag}}(\mathbf{Z}) - \textbf{\text{diag}}(\mathbf{W}\mathbf{Z}) \label{Eq_Computation_passage03}
\end{equation}
and where $\textbf{\text{diag}}(\mathbf{X})$ is a column vector containing the diagonal of matrix $\mathbf{X}$.
The SoF index can therefore be found by applying the following, simple, procedure:
\begin{enumerate}
  \item Compute the $\mathbf{W}$ matrix through Equation (\ref{Eq_W_matrix01}).
  \item Find the matrix $\mathbf{Z}$ from Equation (\ref{Eq:FundamentalMatrix01}).
  \item Compute the column vector $\mathbf{d}$ containing the SoF index of each node with Equation (\ref{Eq_Computation_passage03}).
\end{enumerate}

\section{Experiments} \label{Sec_Experiments}

In this experimental section, the SoF density index is assessed on the identification of dense regions on graphs.
Unlike classical clustering methods, the goal here is not to find an exact partition of the data, but only regions of graphs where the nodes are tightly aggregated, 
suggesting some community-like structure.

\subsection{Datasets} \label{SubSec_Datasets}
The performance of the SoF density index is assessed on ten datasets belonging to four groups: 3-communities, 10-communities, S-Sets and NewsGroup datasets.

The 3-communities (resp. 10-communities) datasets are artificial datasets we built: each one is made of three (resp. ten) clusters, created using gaussian distributions $N(\mu,\sigma)$, $\mu$ being the mean (the center of the cluster) and $\sigma^{2}$ the variance of the data. Each cluster is made of 500 nodes, lying in two dimensions. Three values of $\sigma$ (illustrating various degree of overlapping between the communities) were used to build graphs in the 3-communities case: 0.05, 0.1, 0.5 (the standard deviation is the same in each direction, giving isotropic communities). For the 10-communities datasets, the $\sigma$ values are different in the two space directions, $(x,y)$. These values, called $\sigma_x$ and $\sigma_y$ are reported in Table \ref{Fig:Sigma10C} for two sets : $S_1$ with small overlapping  and $S_2$ with strong overlapping. 

\begin{figure}
  \centering
  \subfigure[$S2$]{\label{fig:S2_1}\includegraphics[scale=0.35]{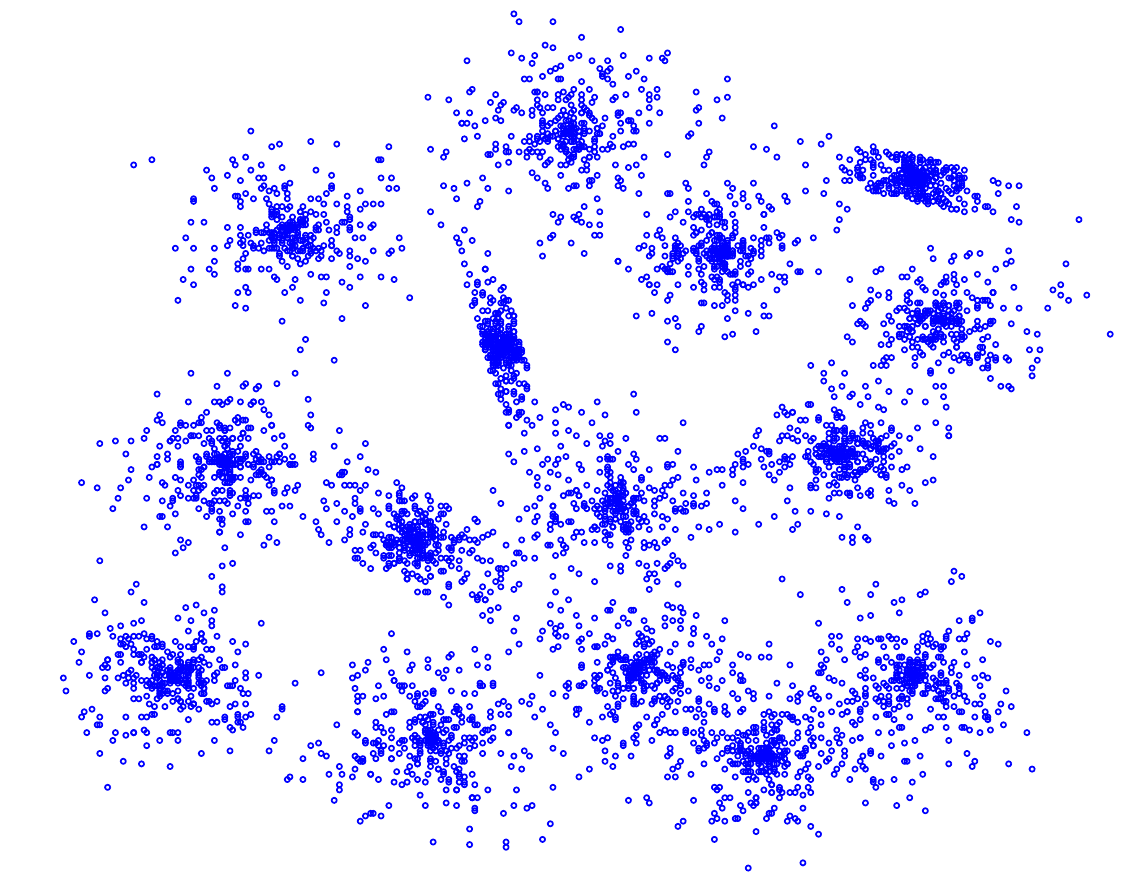}}        
  \subfigure[$S4$]{\label{fig:S4_1}\includegraphics[scale=0.4]{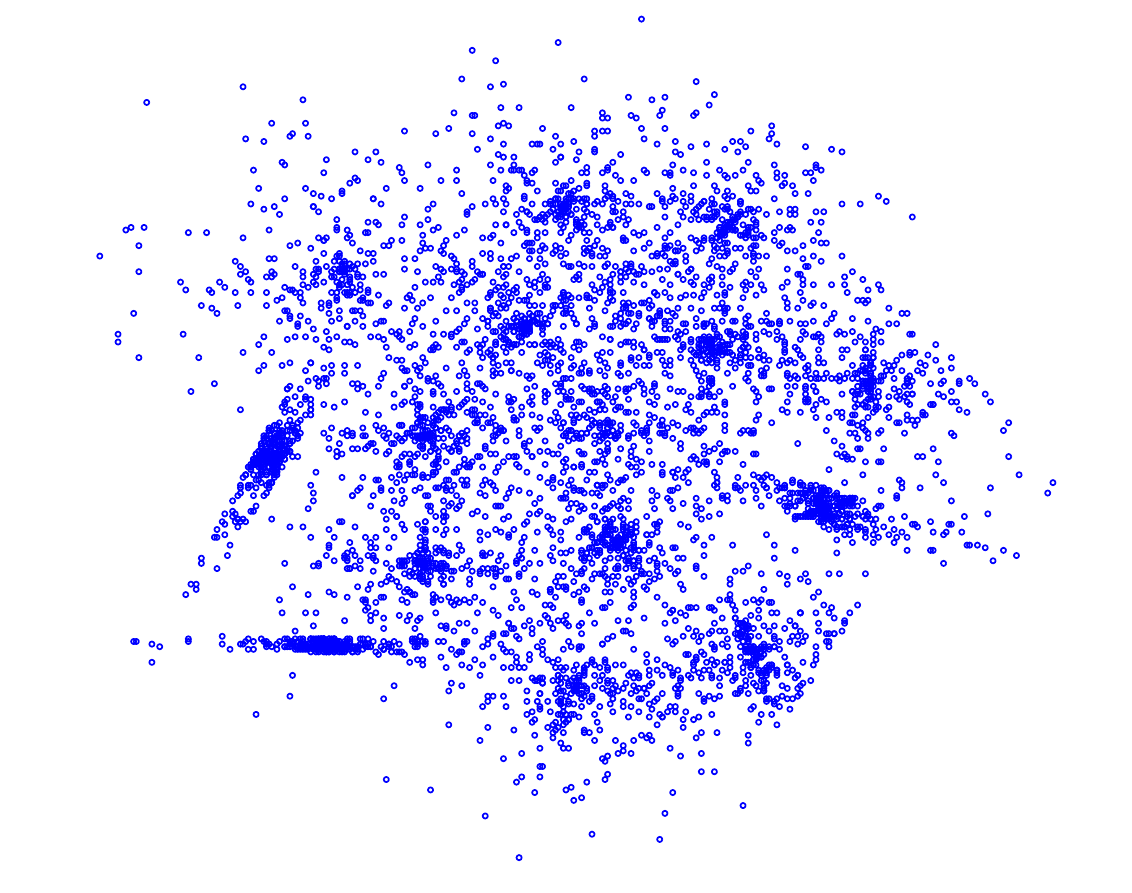}}
  \caption{\textbf{\scriptsize S-Sets datasets.}}
  \label{fig:S-Sets}
\end{figure}

The S-Sets \cite{Franti2006} include two datasets: S2 and S4. They are also based on artificial data and are composed of 5000 two-dimensional observations each, grouped in 15 clusters of various shapes. Figure \ref{fig:S-Sets}  illustrates S2, with well separated clusters and S4, showing more overlapped ones. 

Finally, graphs generated from the Newsgroup dataset are used. This dataset is originally composed of about 20,000 unstructured documents, taken from 20 discussion groups (newsgroups) of the Usernet diffusion list, and composed of 20 classes. For our experiments, three subsets related to different topics are extracted from the original database (NewsGroup1, 2, and 3) \cite{Yen-09}. The graphs of documents were built by sampling at random about 200 documents in each of three classes from three different topics.

\begin{table}
\scriptsize
\begin{center}
\begin{tabular}{|c|c|c|c|c|c|c|c|c|c|c|c|}

   \hline
   S1 &$\, \sigma_{x}$ & 0.8 & 0.5 & 0.5 & 0.8 & 1 & 1 & 0.5 & 0.5 & 0.5 & 1 \\

          &$\, \sigma_{y}$ & 0.8 & 0.5 & 0.5 & 0.5 & 1 & 0.5 & 1 & 1 & 1 & 0.5 \\
   \hline
   S2 &$\, \sigma_{x}$ & 1.8 & 1.5 & 1 & 1.8 & 2 & 1 & 2.5 & 1.5 & 1 & 3  \\

		   & $\, \sigma_{y}$  & 1.6 & 1 & 2.5 & 3 & 2 & 1 & 2 & 2 & 3 & 1.5 \\
   \hline 					
\end{tabular}
\end{center}
\normalsize
\caption{\textbf{\scriptsize $(\sigma_{x},\sigma_{y})$ (standard deviations) values for the 10-communities datasets, for two degrees of overlapping between the clusters (S1 small overlapping, S2 strong overlapping).}}
\label{Fig:Sigma10C}
\end{table}

\subsection{Graph construction} \label{SubSec_Graph Construction}
We constructed the graphs corresponding to the 3/10-communities and the S-Sets datasets using two classical methods: the $\epsilon$-graph and the k-nearest neighbours (k-NN). 

The \textbf{$\epsilon$-graph} computes the euclidean distance between each pair of observations in the dataset and transforms it into an affinity using 
\begin{equation}
	a_{ij} = \exp\left[- \frac{d_{ij}^2}{\sigma^{2}}\right] \label{Eq_Affinities}
\end{equation}
where $d_{ij}$ is the euclidean distance between nodes $i$ and $j$, and $\sigma^{2}$ is the variance of the distances between all the observations in the dataset. 
The nodes are then linked to others only if they show an affinity superior to a certain threshold (80, 90, 95, and 99 percentiles were used). The resulting graphs are undirected, and both the weighted case (where arcs bear the nodes affinities) and the unweighted case are investigated. 

The \textbf{k-NN} graph construction method simply links a node to its k nearest neighbours, i.e., those who have the highest affinity with that node. This relation is not symmetric, giving birth to directed graphs. We transform them into undirected graphs using 
\begin{equation}
	\mathbf{A} \leftarrow \max\left(\mathbf{A,A^T}\right)\label{Symm-KNN}
\end{equation}
where $\mathbf{A}$ is the adjacency matrix of the created graph, and the maximum operator is taken elementwise.

\begin{figure}
  \centering
  \subfigure[NewsGroup1]{\label{fig:News1_1}\includegraphics[scale=0.25]{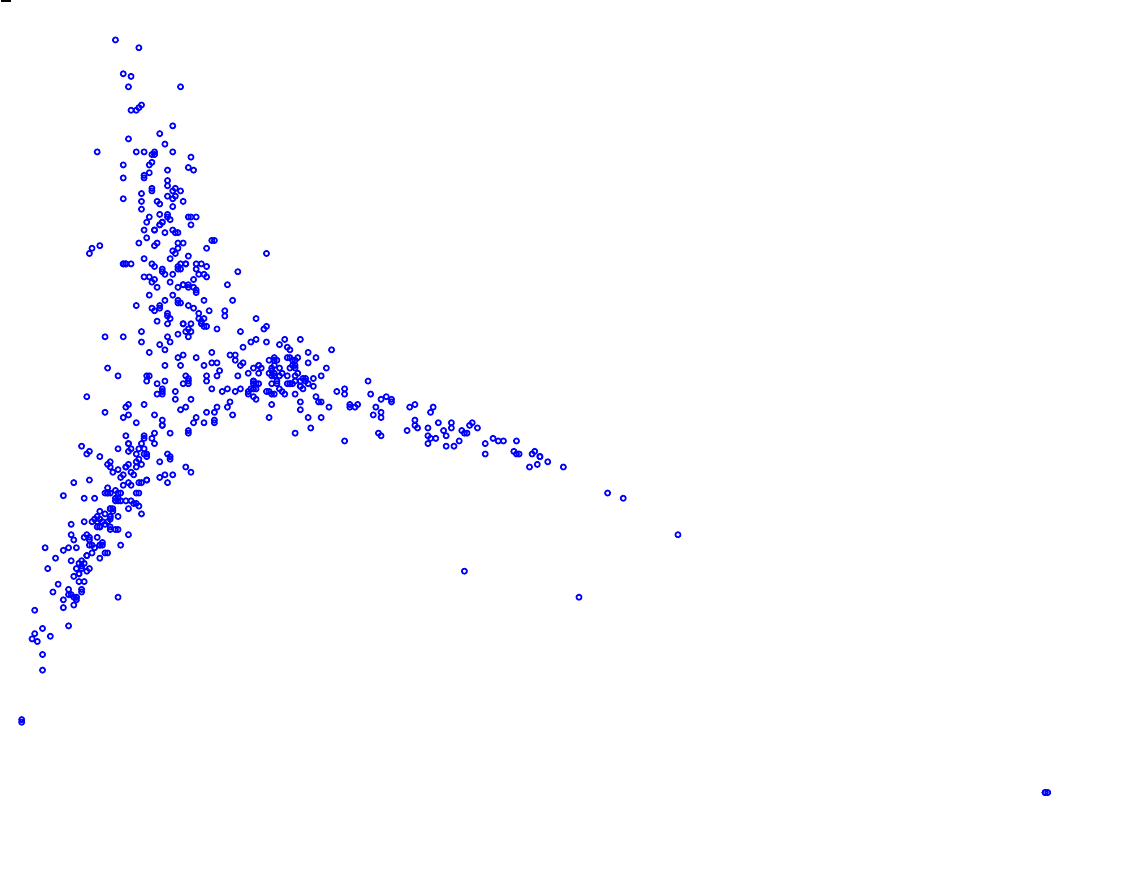}}          
  \subfigure[NewsGroup2]{\label{fig:News2_1}\includegraphics[scale=0.25]{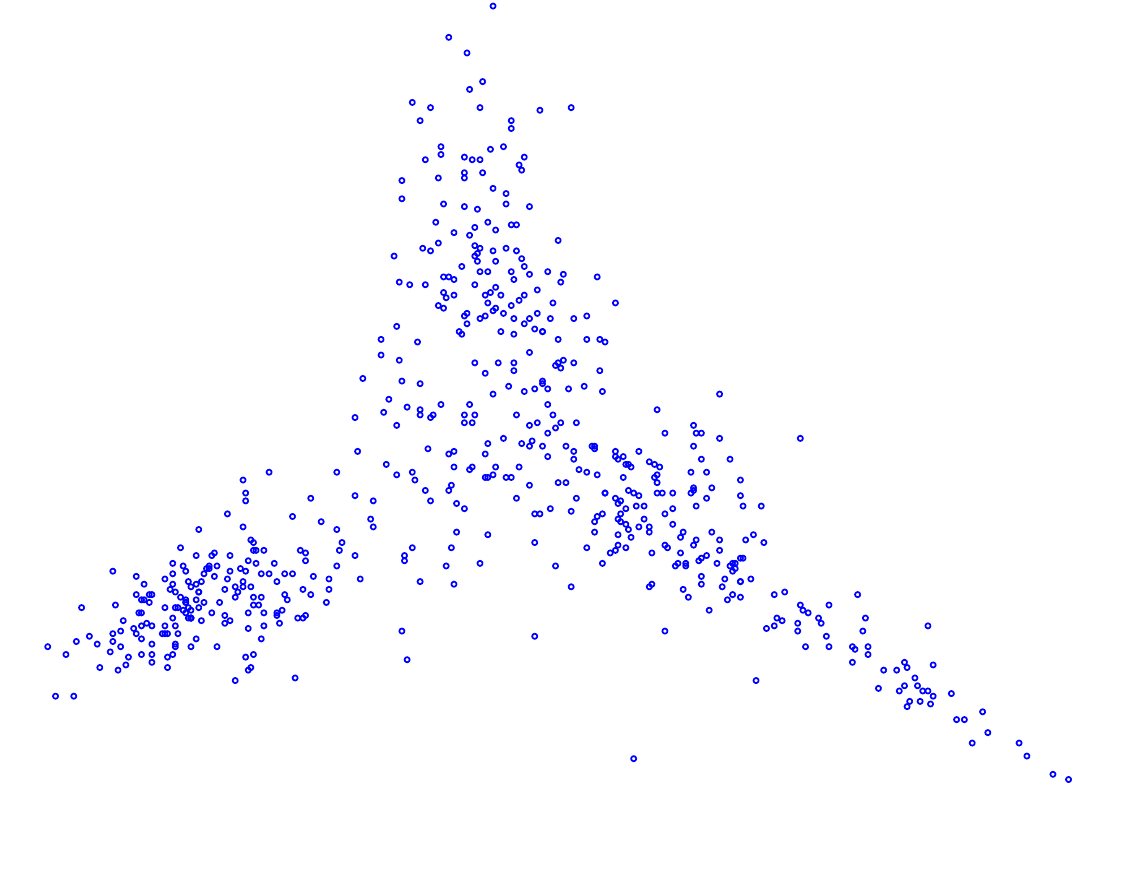}}
  \subfigure[NewsGroup3]{\label{fig:News3_1}\includegraphics[scale=0.25]{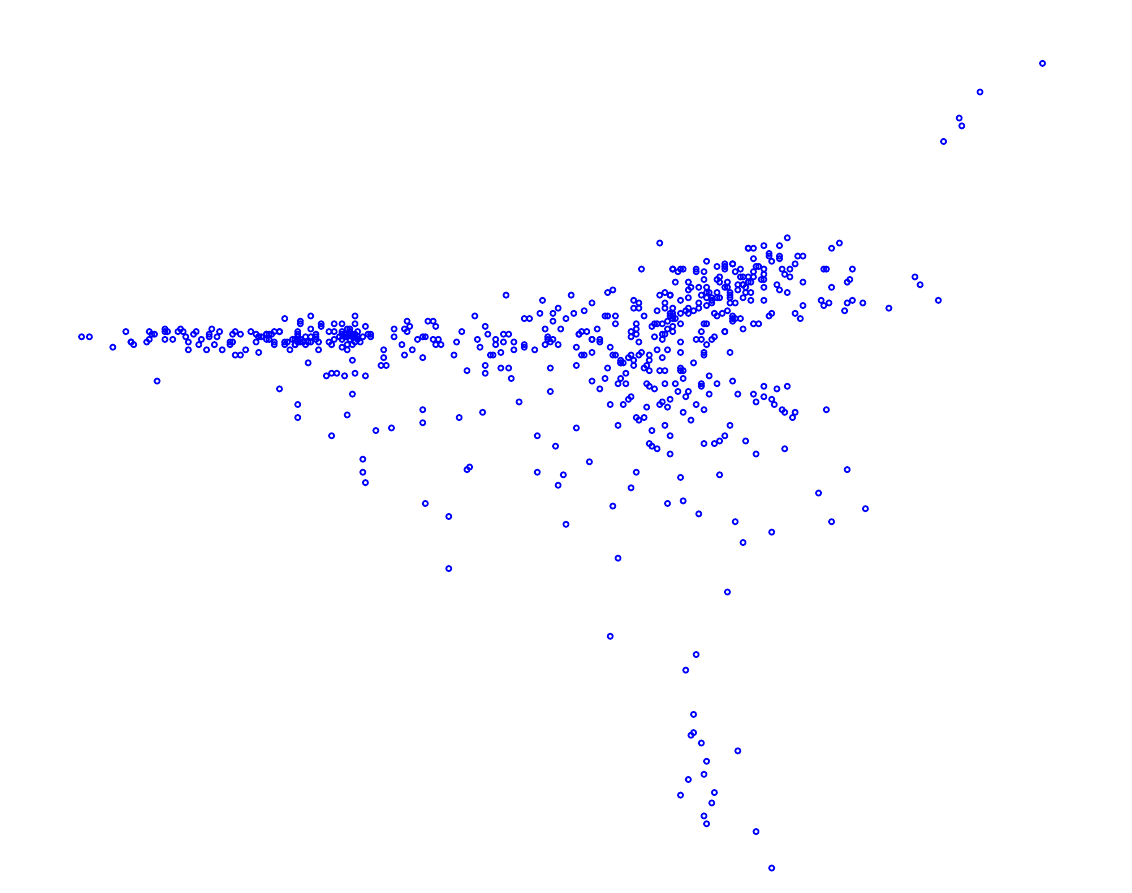}}
  \caption{\textbf{\scriptsize NewsGroup datasets.}}
  \label{fig:News}
\end{figure}

For the NewsGroup datasets, the graphs were already build \cite{Yen-09} and only the adjacency matrices are at our disposal. To visualize those graphs, we use the diffusion maps embedding method \cite{Nadler-2005,Lafon-2006,Yen-2011} in two dimensions (see Figure \ref{fig:News}), whose output is the new spatial coordinates of the nodes. The corresponding graphs are reconstructed with the $\epsilon$-graph method, allowing to compute the density index on the nodes. Indeed, trying to proceed inversely (computing the densities before the diffusion map embedding) is not visually accurate: during the embedding, the nodes are spatially rearranged and the color of the nodes (indicating high or low density, see below) do not reflect the true density of the 2-D embedding.

The cost matrices used in the evaluation of the SoF density index are then computed as the reciprocals of the affinity matrices constructed above.

\subsection{Evaluation methods} \label{SubSec_Evaluation}
We use two methods to evaluate to which extent the high density areas are well identified: Spearman's correlation (only applicable to 3/10-communities datasets) and visual checking (applicable to all datasets).

Firstly, since the probability density function is known for every node of the 3/10-communities datasets (i.e., the exact parameters' values of the gaussian distributions are known), we compute \textbf{Spearman's correlation} between those true densities and the SoF densities. 

Secondly, we perform a \textbf{visual checking} on the graphs by superimposing the density index on the representation of the nodes. This is done by assigning each node a color: from dark blue for nodes presenting a low density value to dark red for nodes presenting a high density value. 

Concerning the tuning of the $\theta$ parameter in the SoF method, we used the correlation method on the 3/10-communities graphs. The parameter's value giving the highest correlation score (for threshold graphs, $\theta=5$ and for k-NN graphs, $\theta=50$) is then used for the 3/10-communities as well as for the S-Sets and Newsgroup datasets. 

The results obtained with the SoF density index are finally compared with two other measures for identifying dense zones: the strength (Str) and the clustering coefficient (CC).

\subsection{Results} \label{SubSec_Results}

\subsubsection*{Correlation results}
\begin{figure}
  \centering
  \subfigure[3-communities - k-NN]{\label{fig:Corr_3comm_NN}\includegraphics[scale=0.55]{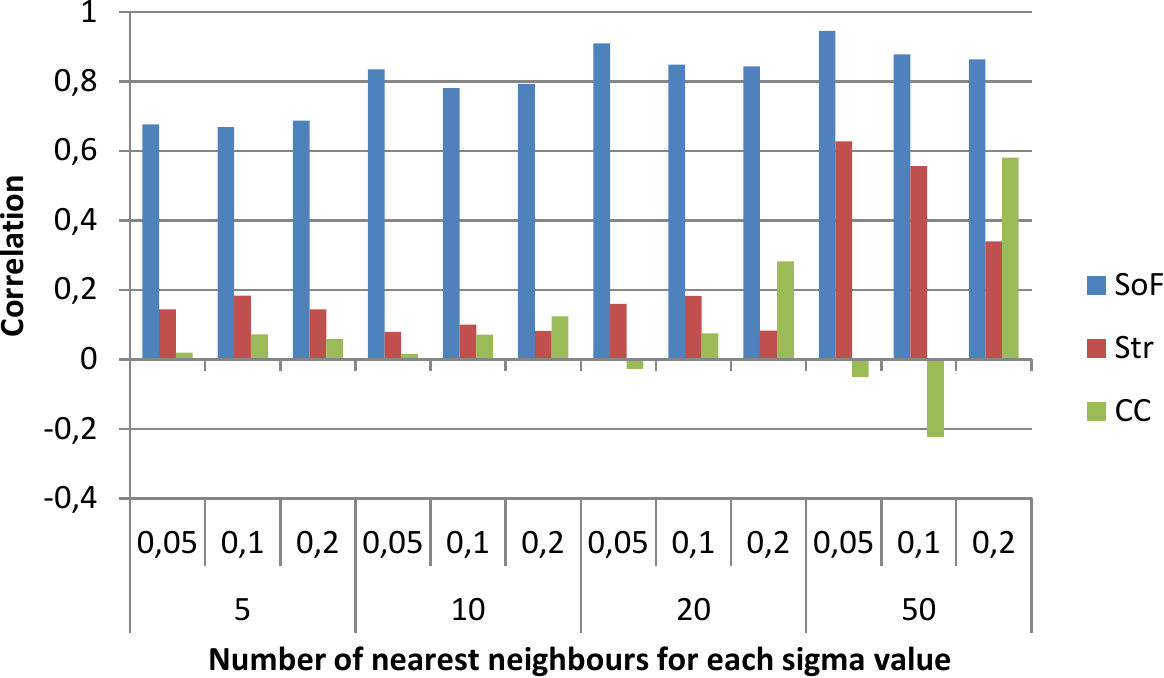}}         
  \subfigure[3-communities - Th - W]{\label{fig:Corr_3comm_Th_W}\includegraphics[scale=0.55]{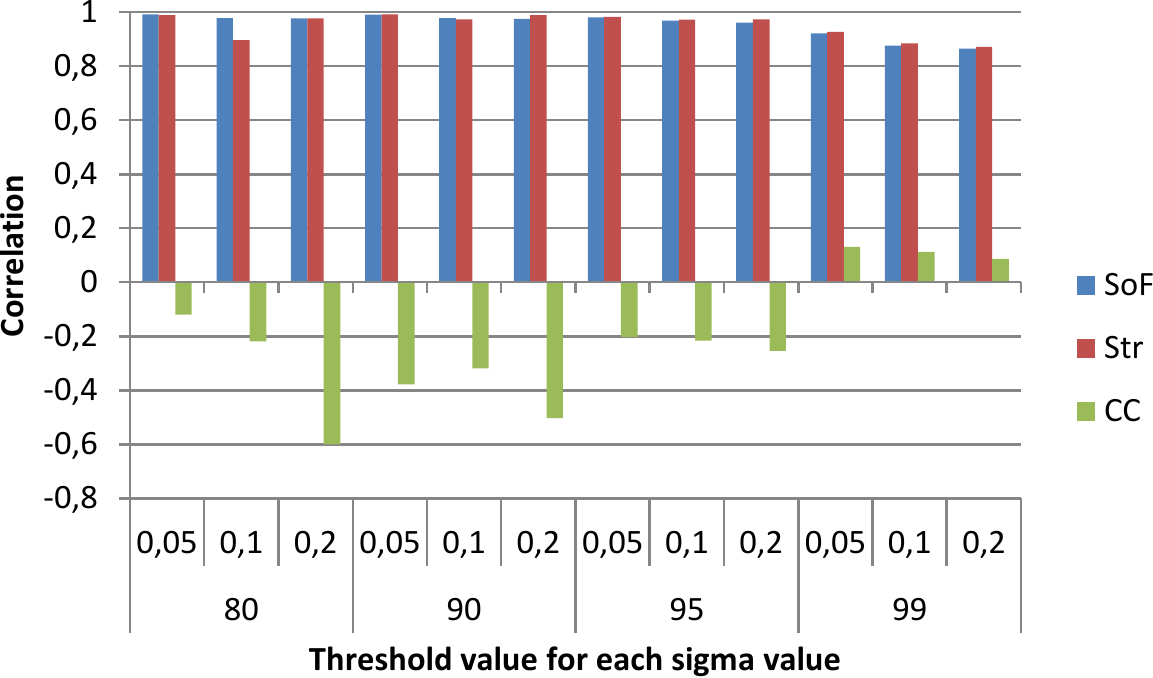}}      
  \subfigure[10-communities - k-NN]{\label{fig:Corr_10comm_NN}\includegraphics[scale=0.55]{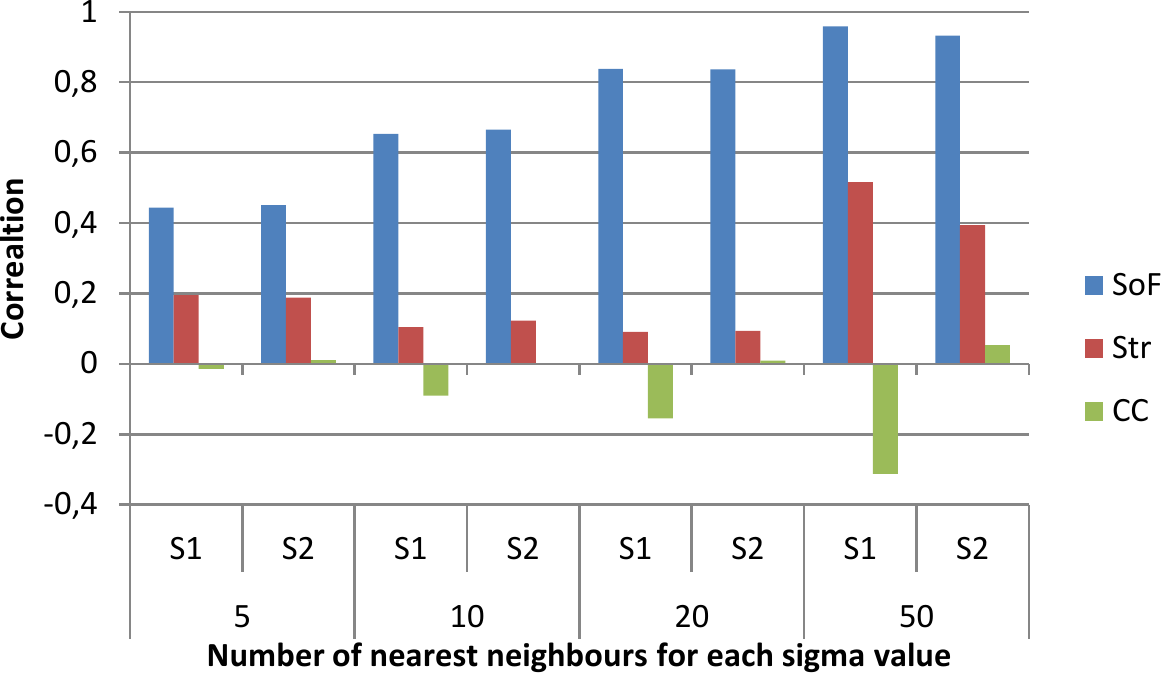}} 
  \subfigure[10-communities - Th - W]{\label{fig:Corr_10comm_Th_W}\includegraphics[scale=0.55]{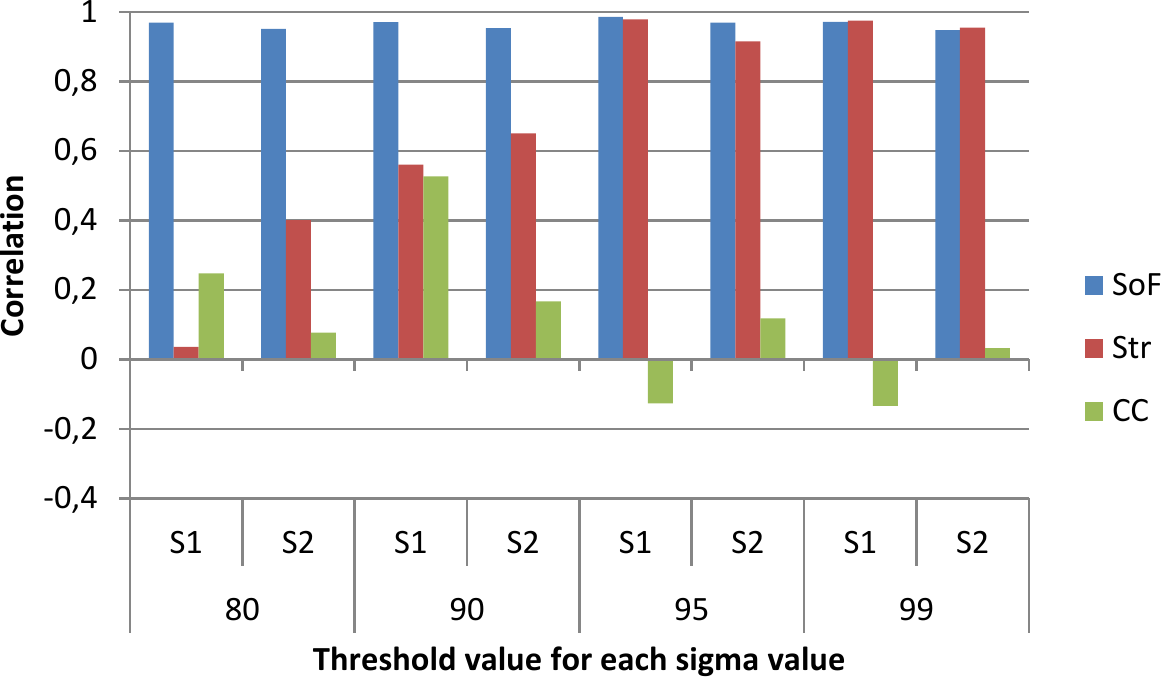}}
  \subfigure[10-communities - Th - U]{\label{fig:Corr_10comm_Th_U}\includegraphics[scale=0.55]{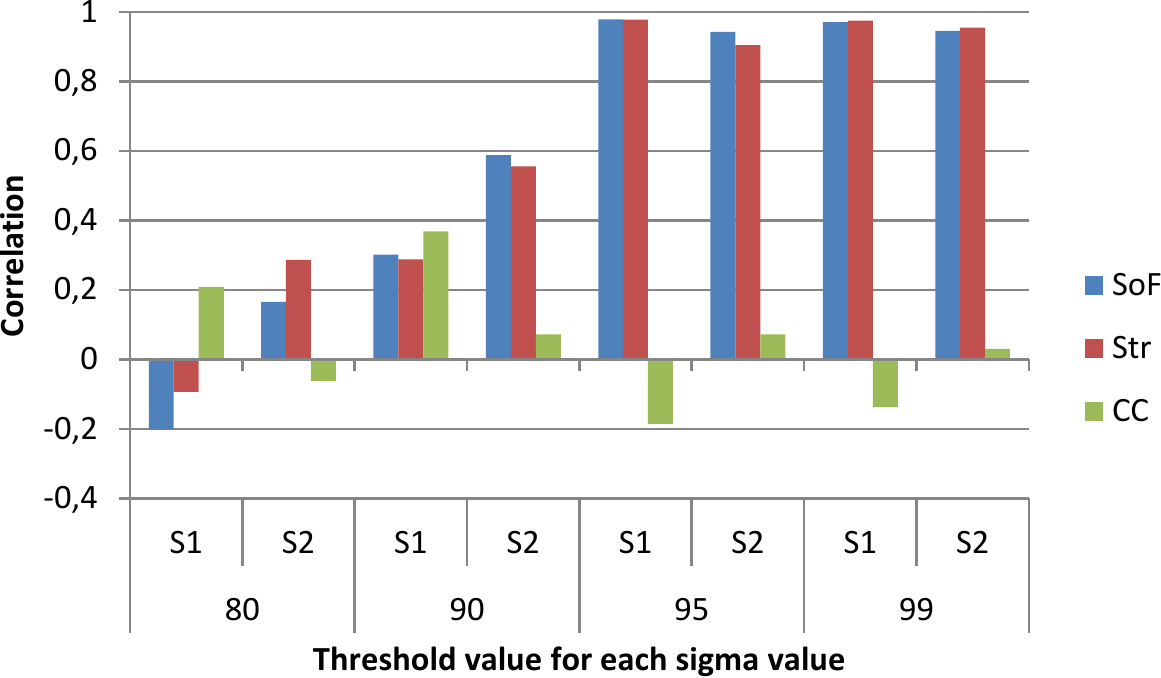}}   
  \caption{\textbf{\scriptsize Correlation results for the 3/10-communities datasets, between true density and SoF density index, Strength, CC. Weighted (W) and Unweighted (U) graphs are considered, constructed with k-Nearest Neighbours (k-NN) or Threshold (Th) methods.}}
  \label{fig:Corr}
\end{figure}

The correlation results for 3/10-communities are displayed in Figure \ref{fig:Corr}. 

When using the k-NN for constructing graphs, the SoF density index is clearly superior to the strength and to the clustering coefficient (the latter performs badly in every situation, and is not further considered in the sequel of this section). This may be explained by the fact that the information concerning the connectivity is useless in this case, as all the nodes have theoretically almost the same degree. The SoF density index then makes a better use of the affinities borne by the arcs of the graphs than the strength does, which explains its superior results. 

When using the $\epsilon$-graph construction method, the results are not so clear. The results obtained by the strength and the SoF index for the 3-communities case have a correlation with the true density of almost one and are practically identical  (only the weighted case is represented here in Figure \ref{fig:Corr_3comm_Th_W}, as the unweigthed case gives similar results). The SoF index is clearly better on the 10-communities datasets in the weighted case and for low threshold ($\epsilon$) values. Indeed, the number of neighbours increases dramatically when the threshold is low, and again the information of connectivity becomes useless, nodes having each a large degree value. The only useful information are the affinities and, as before, the SoF density index uses it in a more efficient way than the strength. When the threshold is higher the two measures converge to the same value. In the unweighted case (no affinity information), the SoF index and the strength behave similarly: the correlations are low for low threshold values and increase when the threshold increase. The good results of both the SoF index and the strength in the 3-communities dataset with threshold and unweighted arcs are probably due to the fact that these datasets are quite smaller and simpler to handle, having only 3 clusters instead of 10, distributed with gaussians of the same variance in all directions. 

A first conclusion can de drawn so far: the SoF density index is much more stable and independent from the type of graph than the strength (and the clustering coefficient).

\subsubsection*{Visual results}
\begin{figure}
  \centering
  \subfigure[$\sigma=0.05$]{\label{fig:ArtGraphs005_2}\includegraphics[scale=0.18]{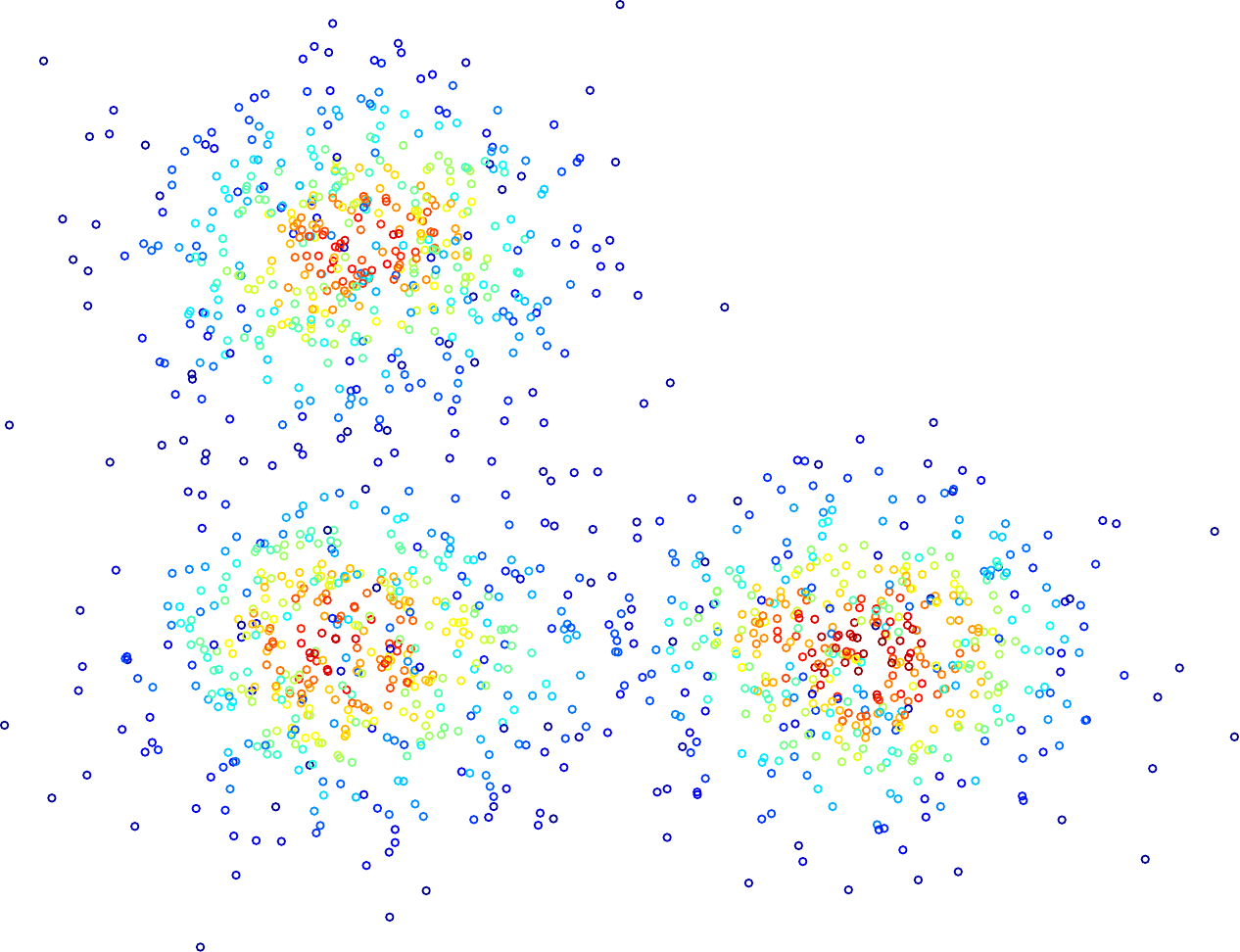}}           
  \subfigure[$\sigma=0.1$]{\label{fig:ArtGraphs01_2}\includegraphics[scale=0.18]{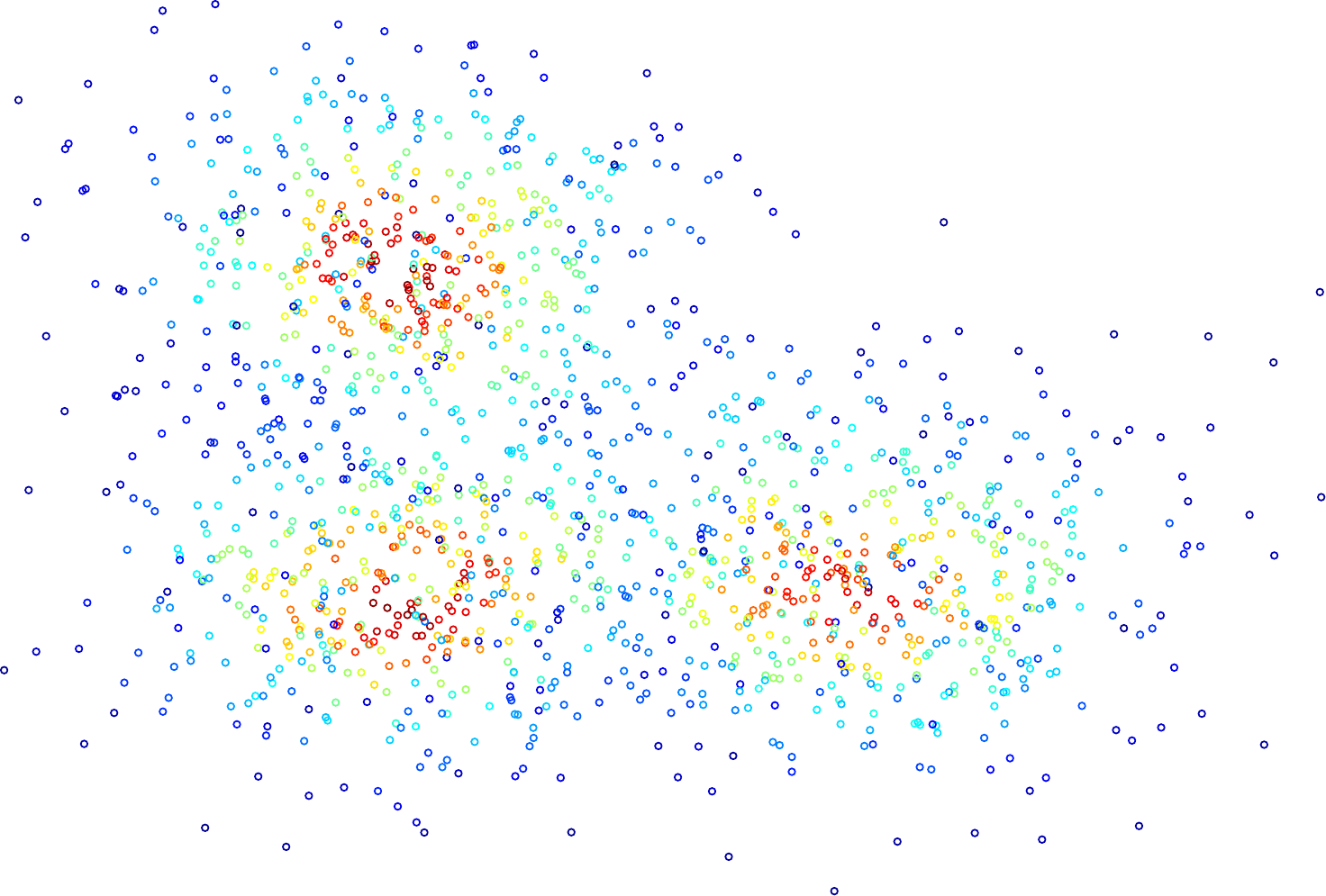}}
  \subfigure[$\sigma=0.5$]{\label{fig:ArtGraphs05_2}\includegraphics[scale=0.18]{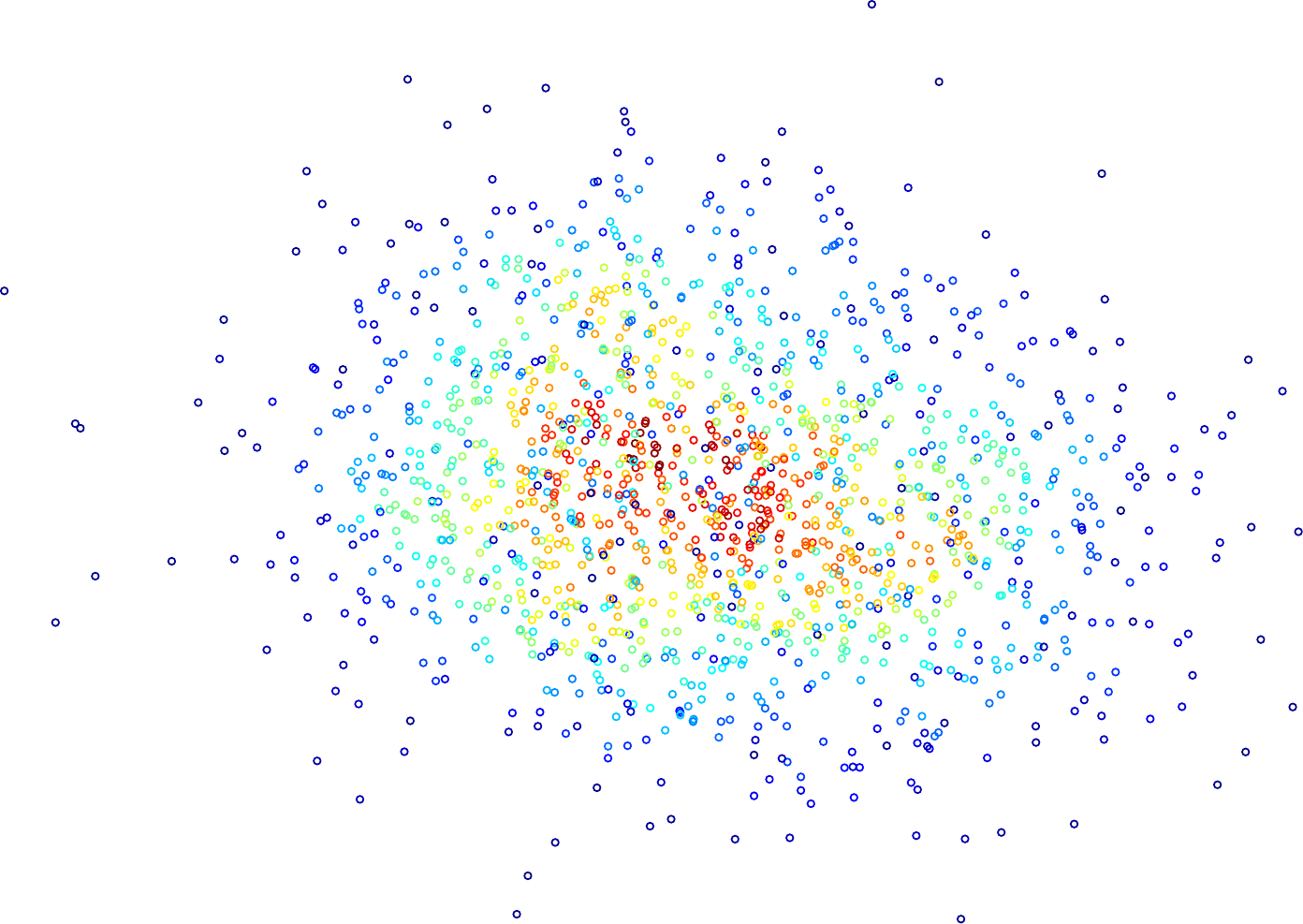}}
  \caption{\textbf{\scriptsize 3-communities datasets for various $\sigma$ values with the SoF density index superimposed.}}
  \label{fig:ArtGraphs_SOF}
\end{figure}

\begin{figure}
  \centering
  \subfigure[True density]{\label{fig:10CTrueD}\includegraphics[scale=0.4]{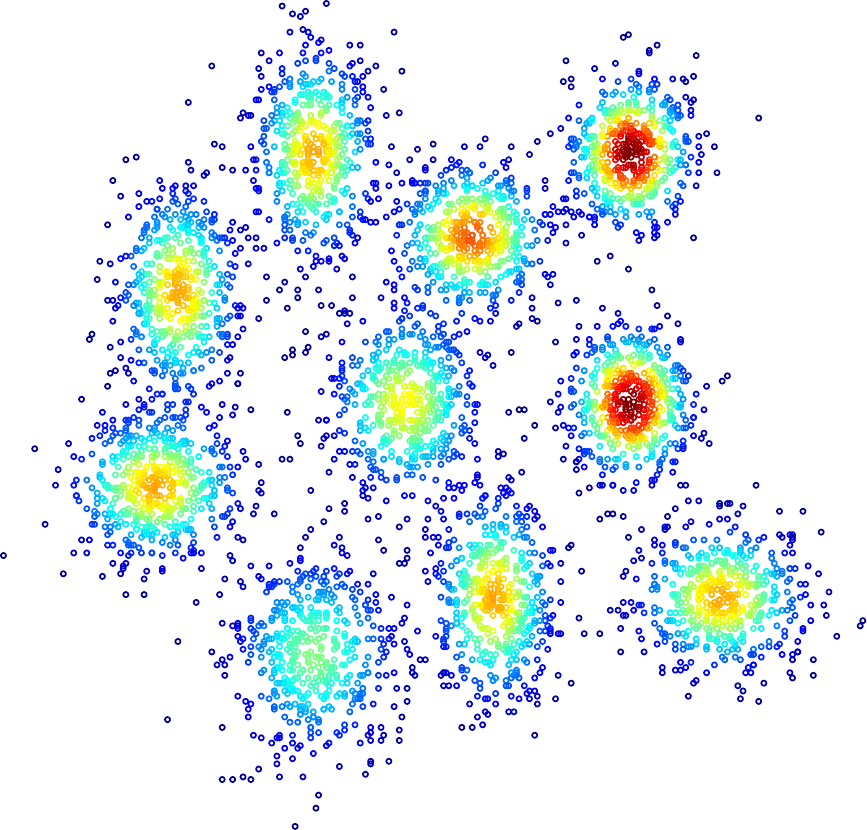}}        
  \subfigure[SoF]{\label{fig:10CSOF}\includegraphics[scale=0.4]{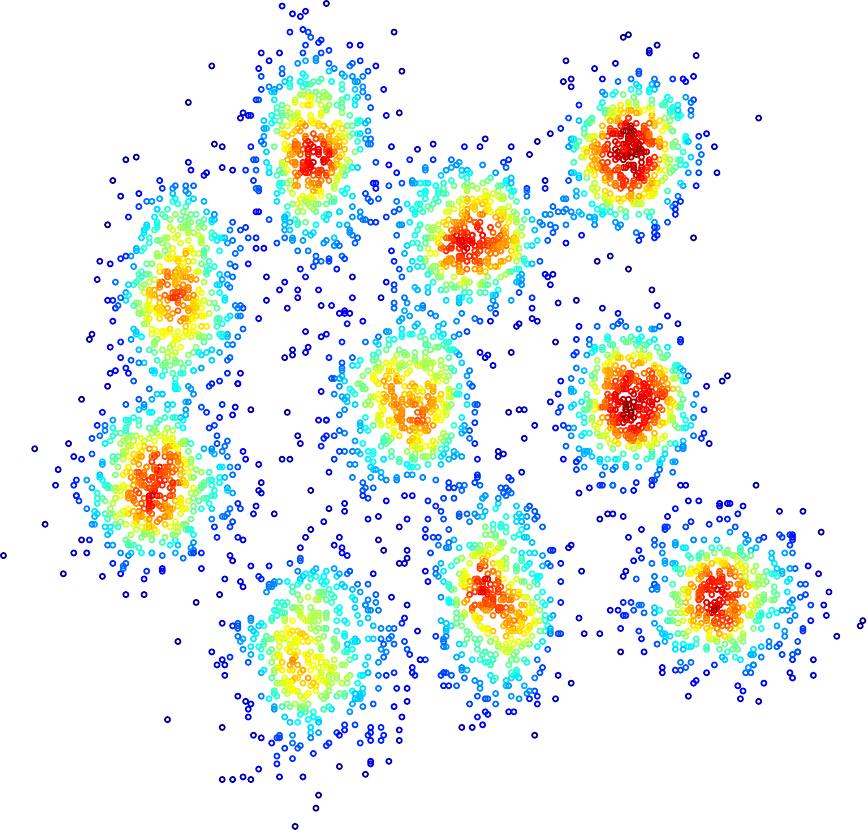}}
  \caption{\textbf{\scriptsize 10-communities dataset (low sigma values S1) with true density (left figure) and SoF density index (right figure) superimposed.}}
  \label{fig:10Comm}
\end{figure}

The visual results confirm the correlation results described above. As there are many different cases for the 3/10-communities datasets, only few visual examples, representative of the overall behavior of the density measures, are shown. For instance, Figure \ref{fig:ArtGraphs_SOF} shows the 3-communities datasets (threshold 95, weighted) with the SoF density index superimposed. It can be observed on this simple example that the SoF density index is able to recover the dense areas of the clusters. In the 10-communities datasets, a visual confirmation is given in Figure \ref{fig:10Comm}. The SoF density index is visually very close to the true density and the highly dense regions are well identified.

\begin{figure}
  \centering
  \subfigure[$S2$]{\label{fig:S2_2}\includegraphics[scale=0.35]{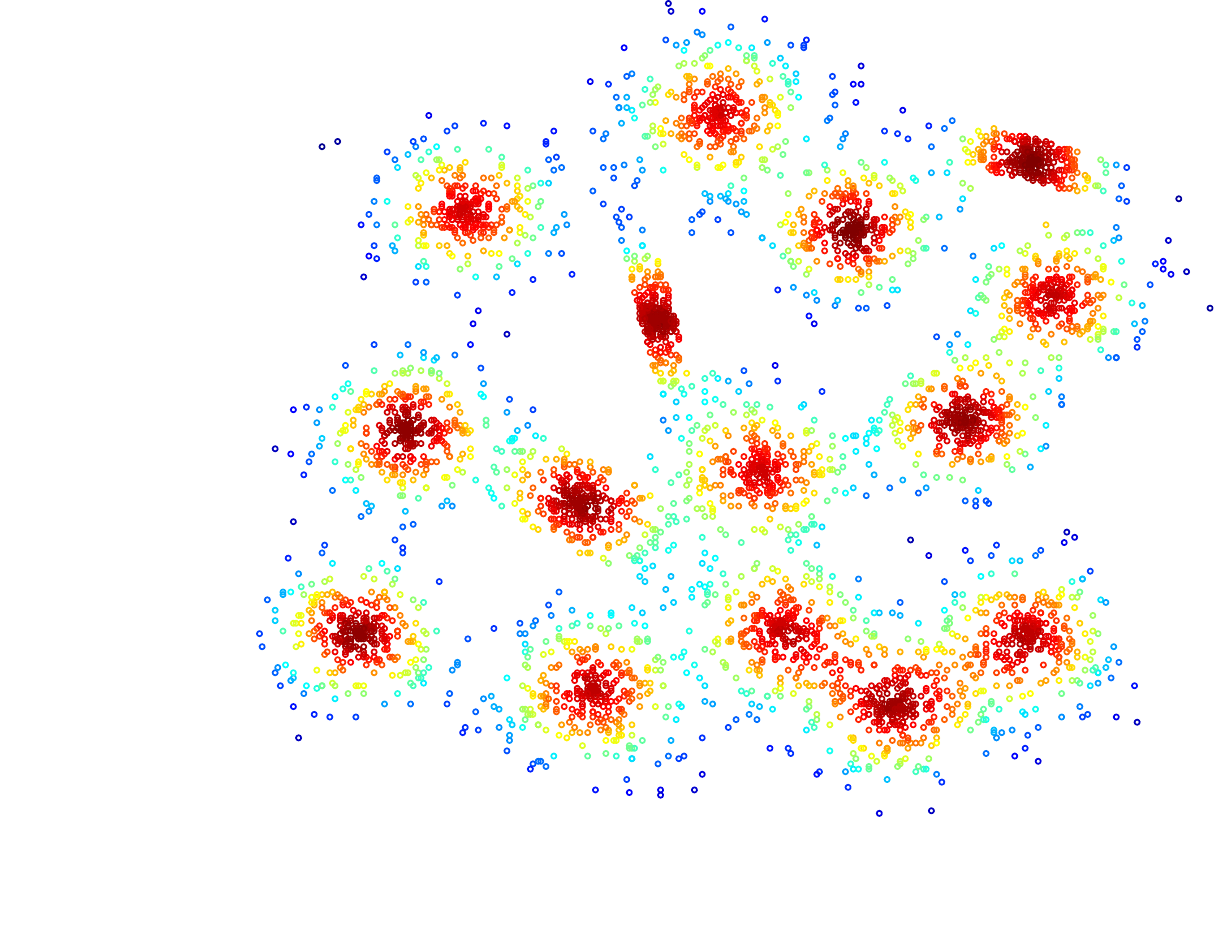}}  
  \subfigure[$S4$]{\label{fig:S4_2}\includegraphics[scale=0.38]{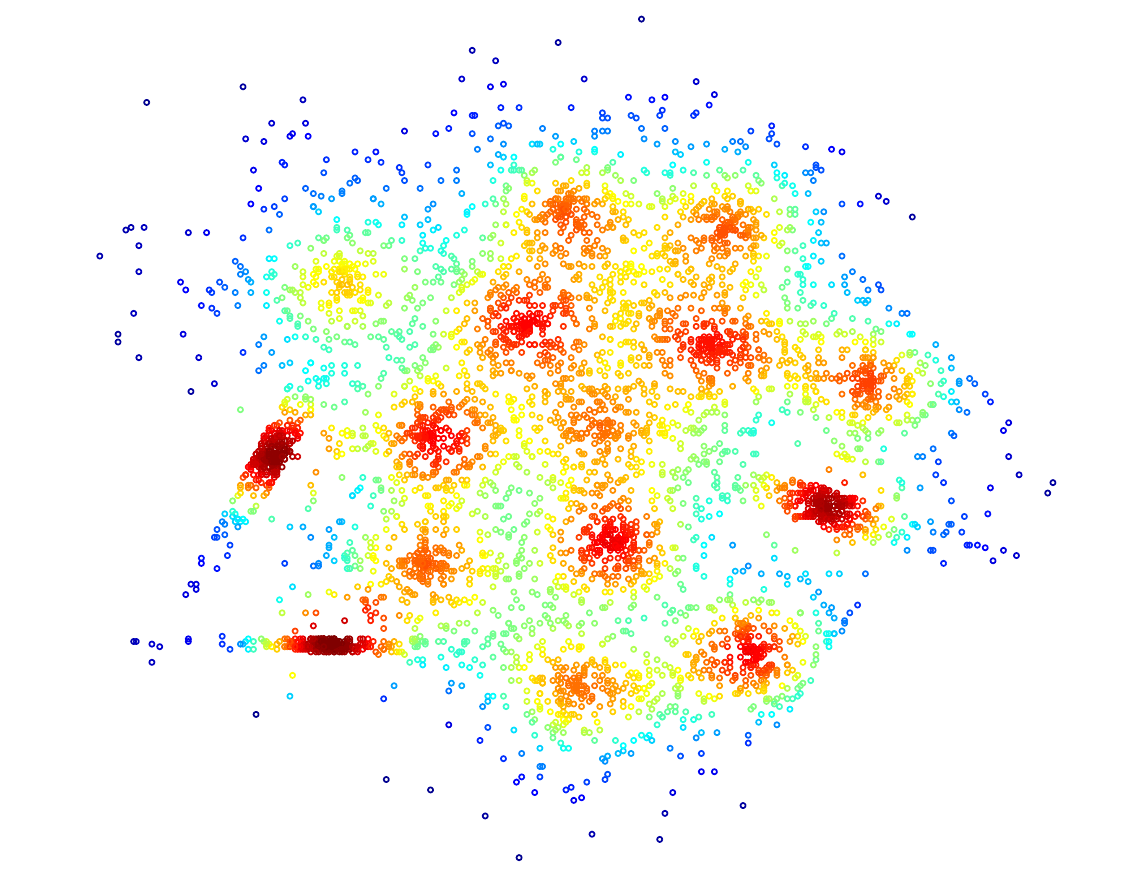}}
  \caption{\textbf{\scriptsize S-Sets datasets with SoF density index superimposed (Threshold 95, Weighted).}}
  \label{fig:S-Sets_SOF}
\end{figure}

\begin{figure}
  \centering
  \subfigure[$S2$]{\label{fig:S2_3}\includegraphics[scale=0.35]{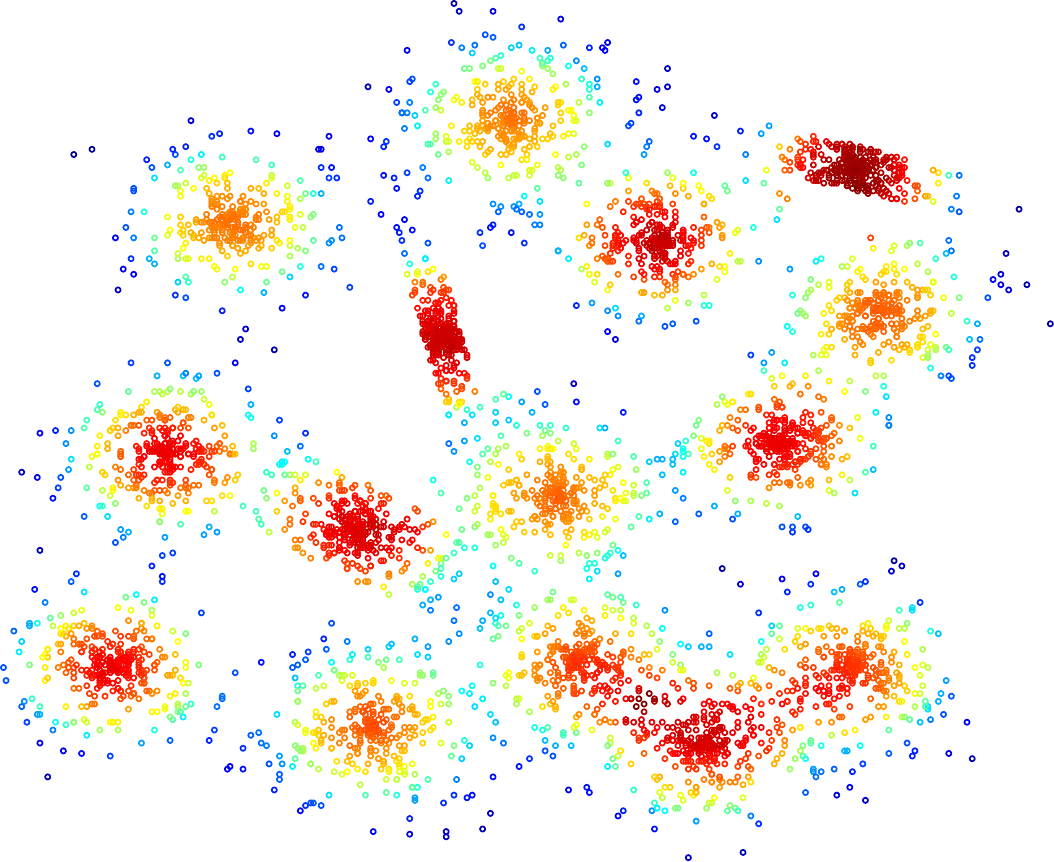}}  
  \subfigure[$S4$]{\label{fig:S4_3}\includegraphics[scale=0.38]{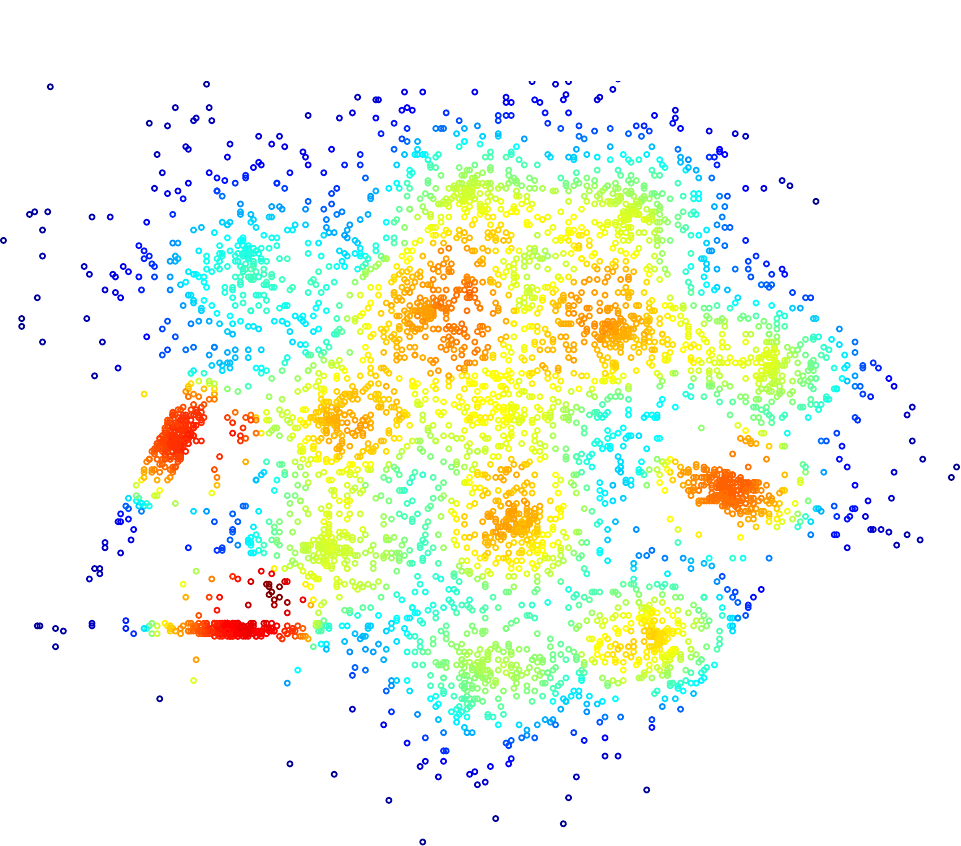}}
  \caption{\textbf{\scriptsize S-Sets datasets with strength superimposed (Threshold 95, Weighted).}}
  \label{fig:S-Sets_Strength}
\end{figure}

\begin{figure}
  \centering
  \subfigure[$S2$]{\label{fig:S2_4}\includegraphics[scale=0.35]{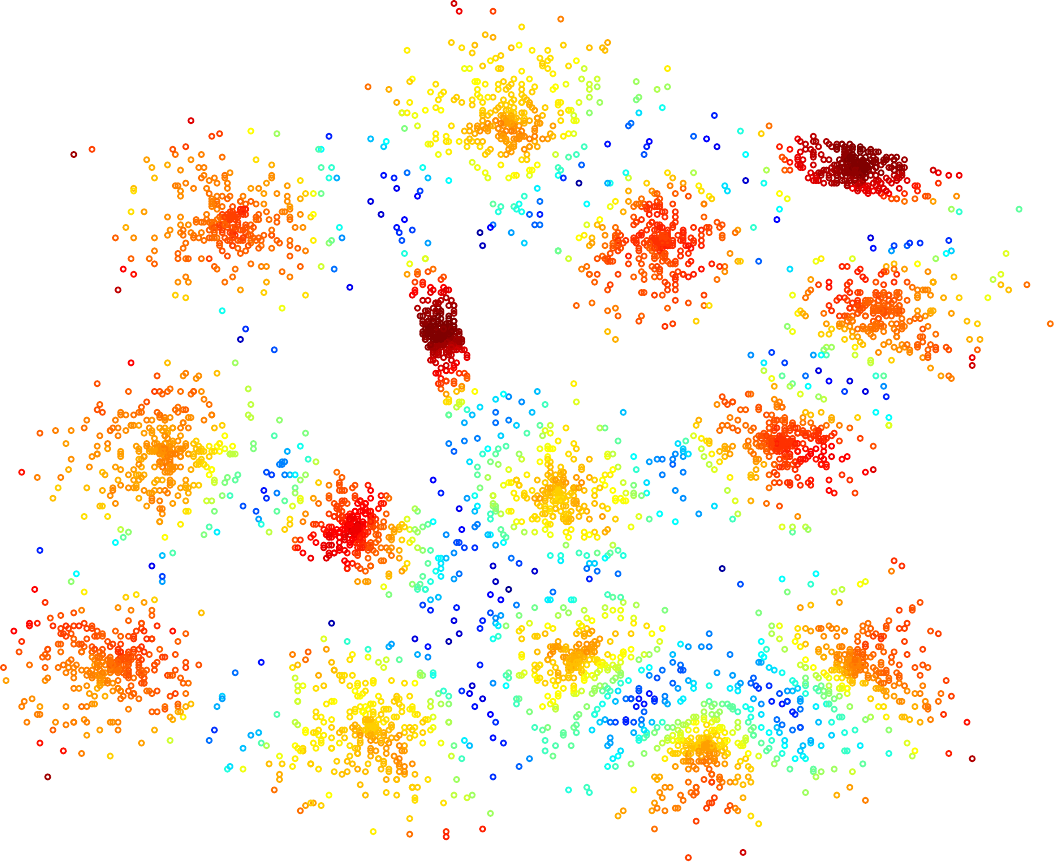}}  
  \subfigure[$S4$]{\label{fig:S4_4}\includegraphics[scale=0.38]{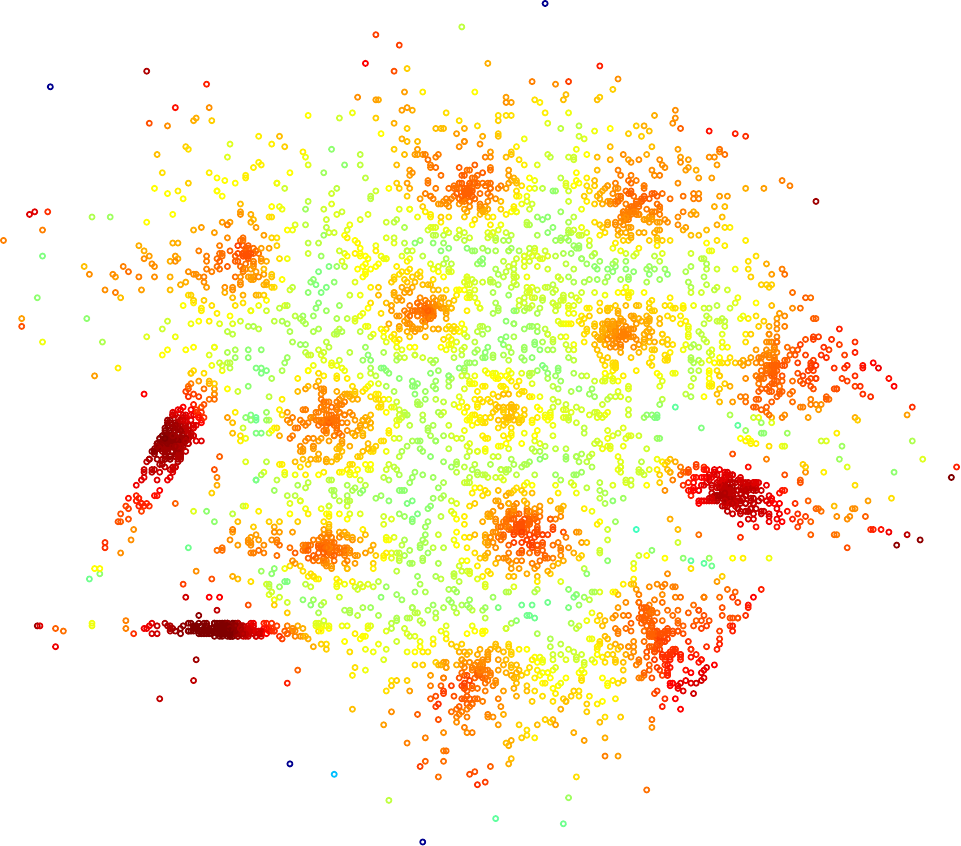}}
  \caption{\textbf{\scriptsize S-Sets datasets with CC superimposed (Threshold 95, Weighted).}}
  \label{fig:S-Sets_CC}
\end{figure}

\begin{figure}
  \centering
  \subfigure[NewsGroup1]{\label{fig:News1_2}\includegraphics[scale=0.2]{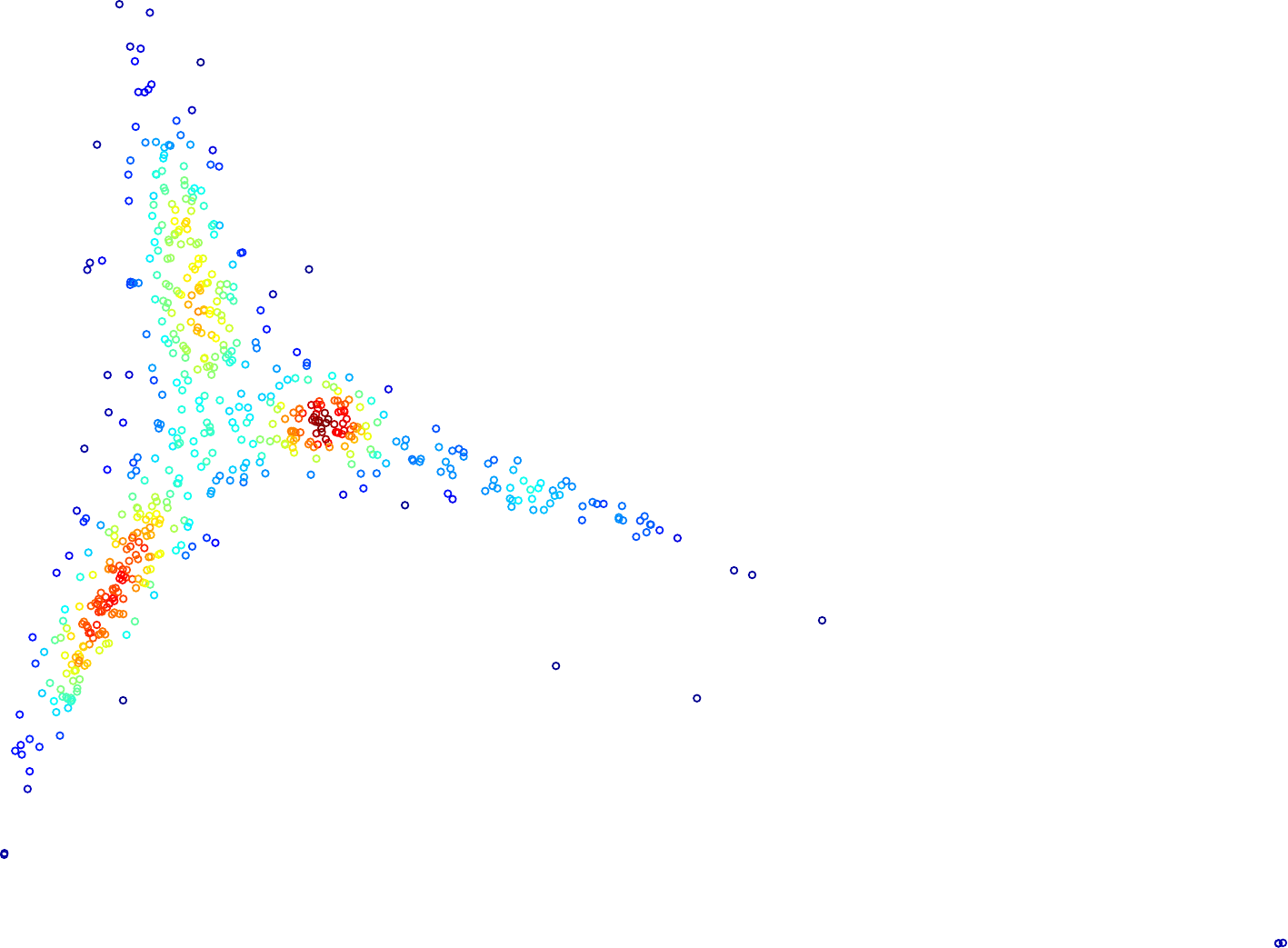}}       
  \subfigure[NewsGroup2]{\label{fig:News2_2}\includegraphics[scale=0.2]{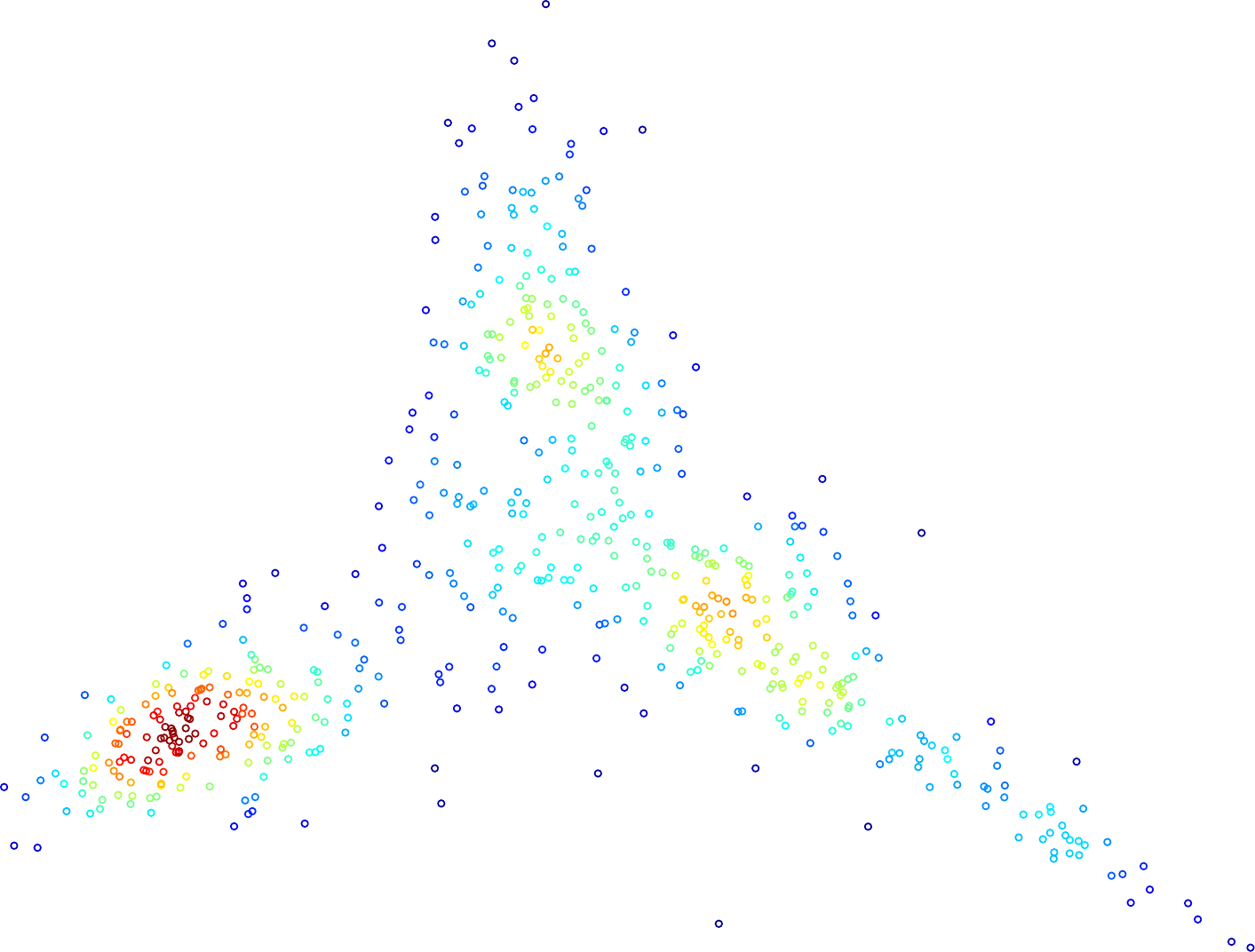}}
  \subfigure[NewsGroup3]{\label{fig:News3_2}\includegraphics[scale=0.2]{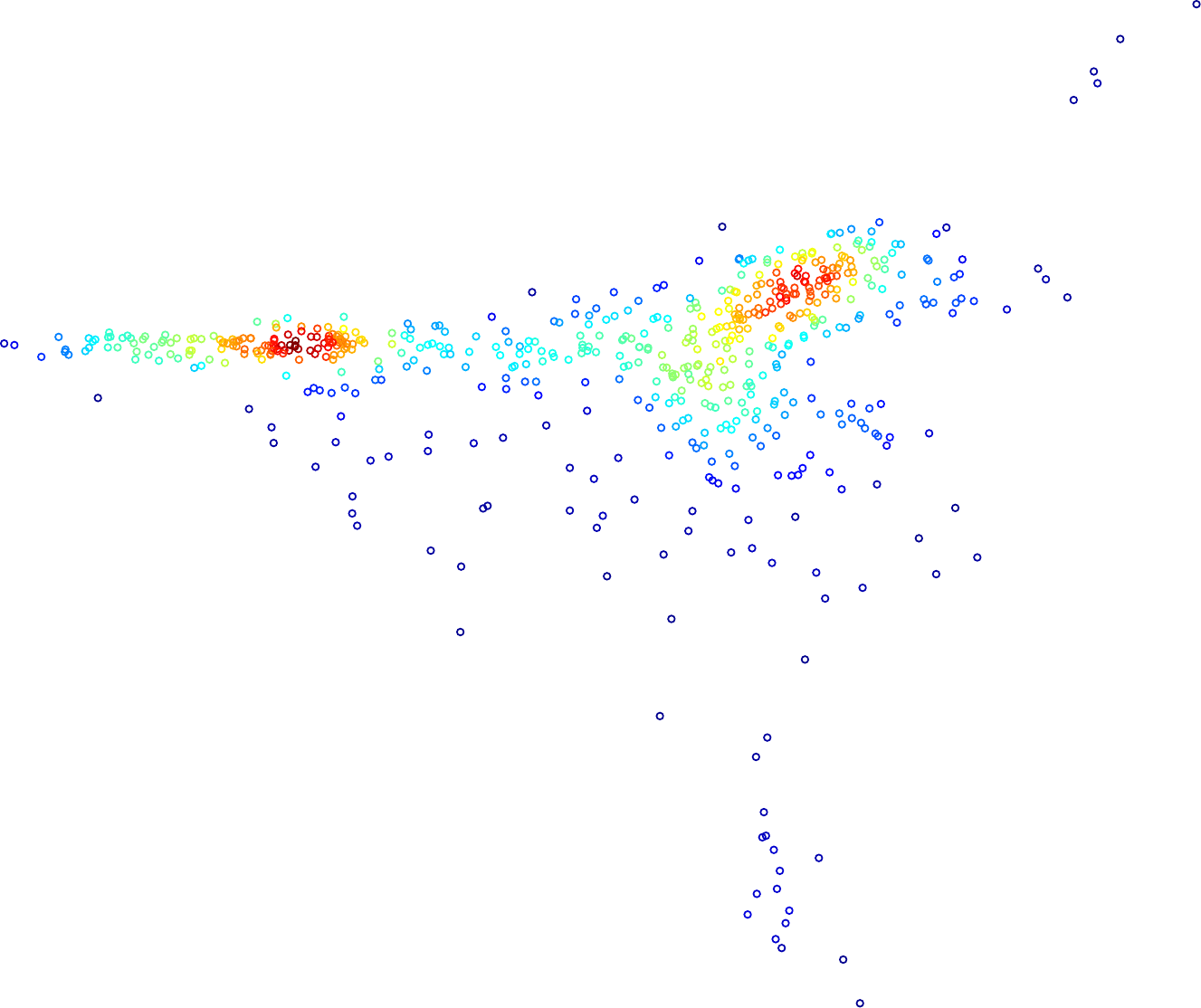}}
  \caption{\textbf{\scriptsize NewsGroup datasets with SoF density index superimposed (Threshold 95, Weighted).}}
  \label{fig:News_SOF}
\end{figure}

The dense regions on the S-Sets are also well identified. Figure \ref{fig:S-Sets_SOF} shows that the SoF density index is able to recover the 15 densest areas on the S2 and S4 graphs, even if they are tightly aggregated. Str and CC are illustrated on Figures \ref{fig:S-Sets_Strength} and \ref{fig:S-Sets_CC}, showing poor results, mainly on S4. The Newsgroup datasets confirm the results obtained so far (Figure \ref{fig:News_SOF}). For SoF density index and Str, the three clusters are recovered on those graphs, except on NewsGroup3 where two clusters are too tightly intrictated to be differentiated. The CC does not identify correctly the dense areas, like in the 3-communities case.  Figures concerning S-Sets and Newgroup graphs show only results obtained for weighted graphs, as those results are essentially identical for unweighted graphs.

\begin{figure}
  \centering
  \subfigure[$\sigma=0.05$]{\label{fig:ArtGraphs005_1}\includegraphics[scale=0.18]{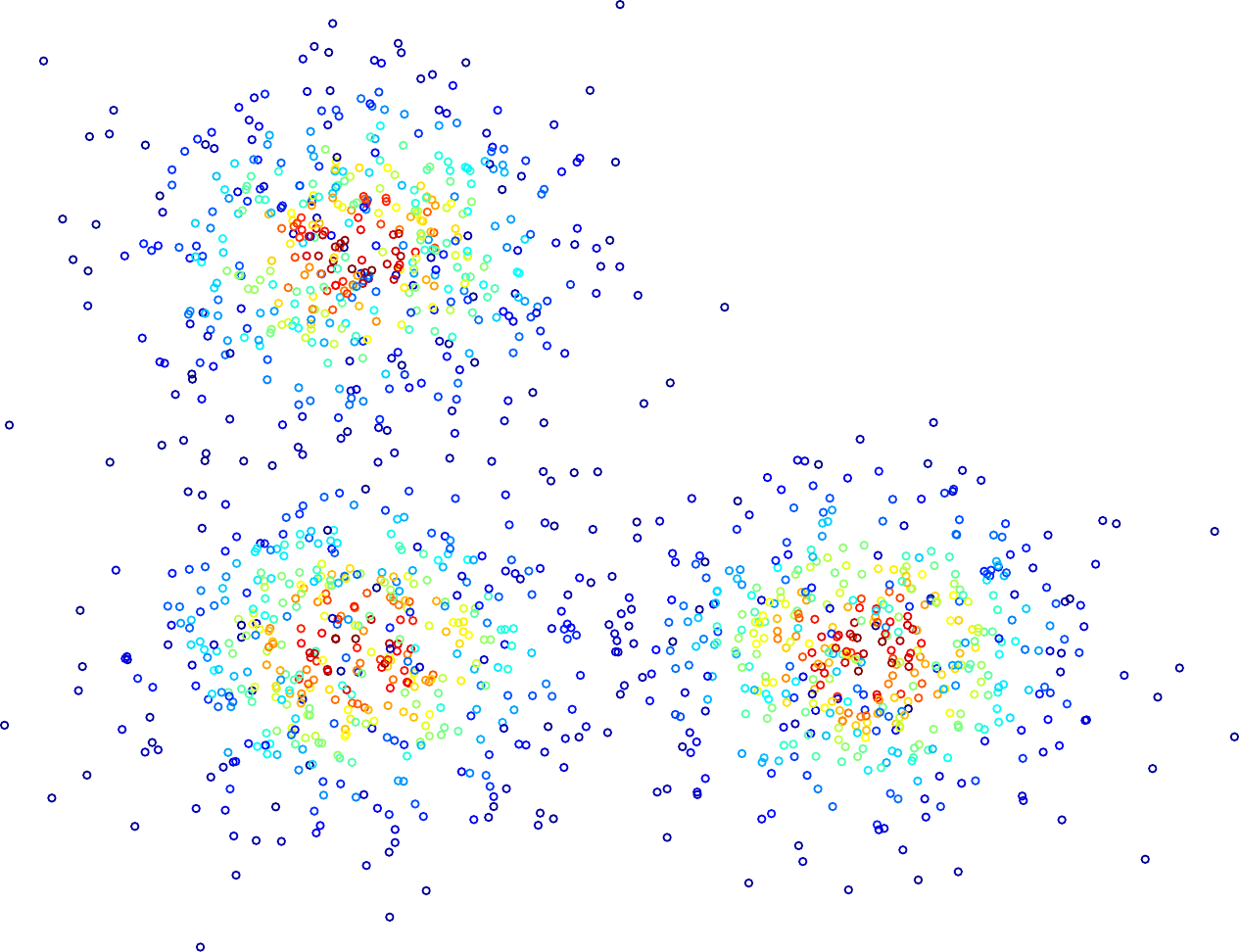}}          
  \subfigure[$\sigma=0.1$]{\label{fig:ArtGraphs01_1}\includegraphics[scale=0.18]{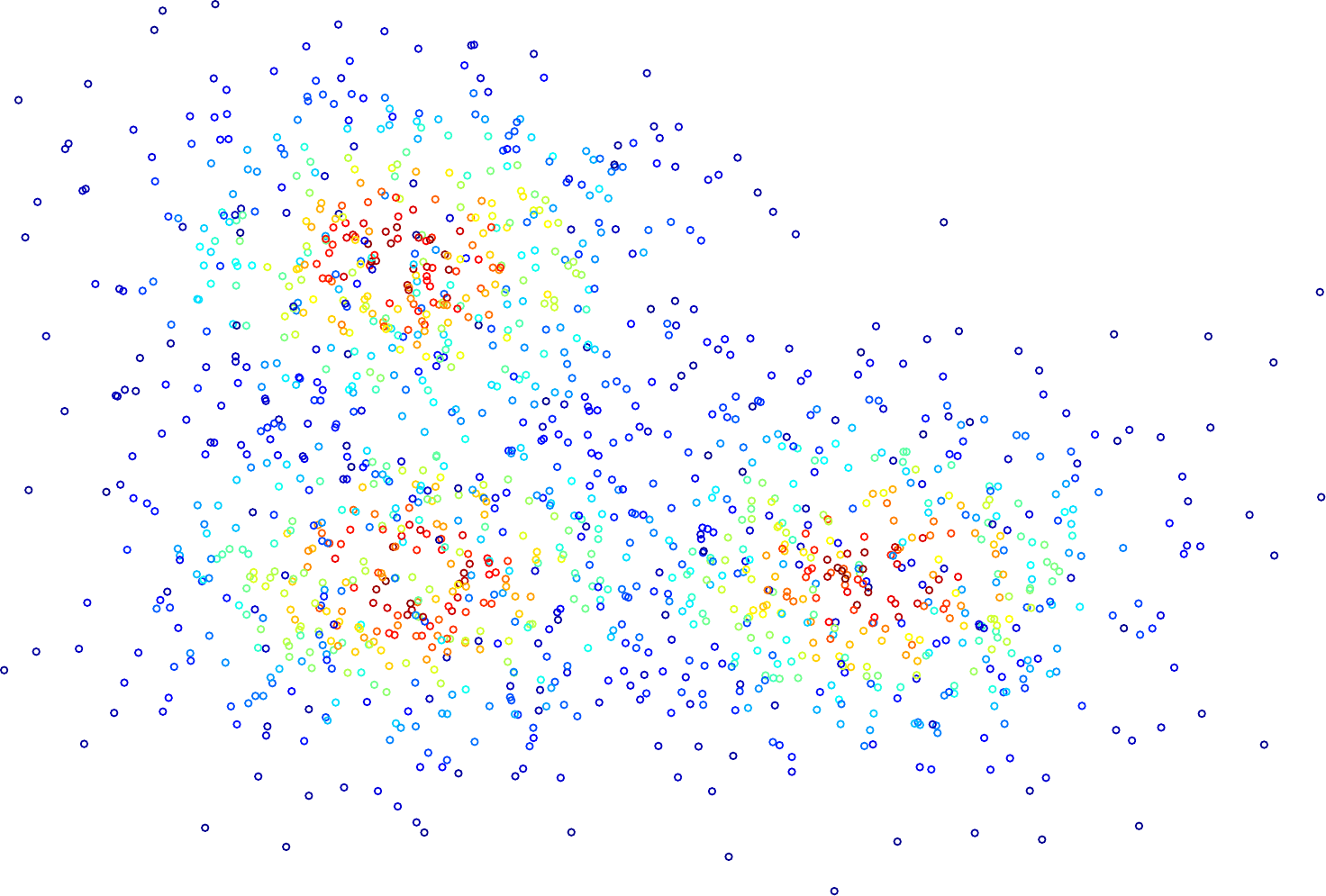}}
  \subfigure[$\sigma=0.5$]{\label{fig:ArtGraphs05_1}\includegraphics[scale=0.18]{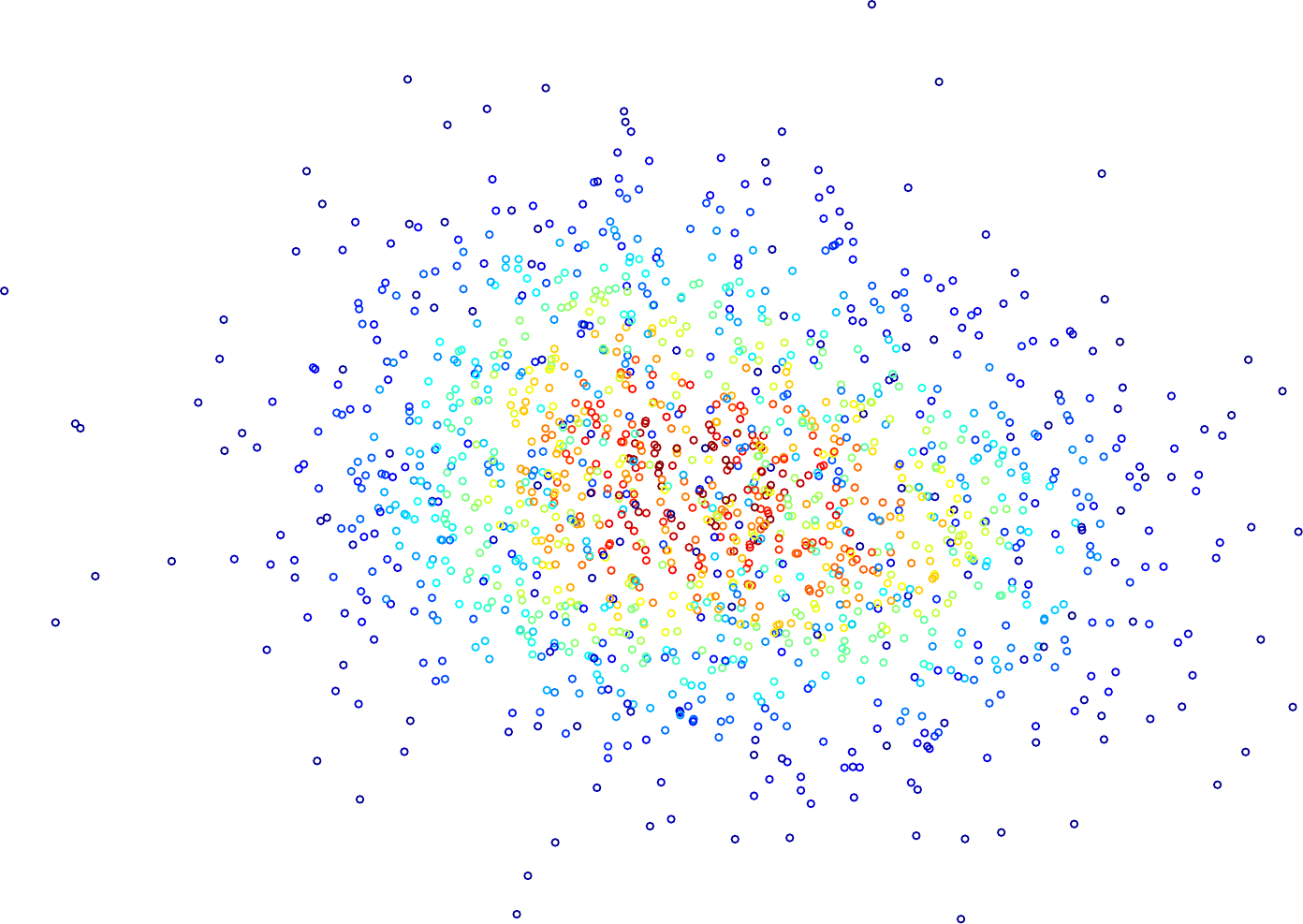}}
  \caption{\textbf{\scriptsize 3-communities datasets for various $\sigma$ values with true density superimposed.}}
  \label{fig:ArtGraphs}
\end{figure}

\section{Conclusion and Perspectives}

This work introduced a new density index on the nodes of a graph. The main idea behind the model is that a node has a high density index if it is present on a large number of (preferably low-cost) forests, together with a high outdegree. This model depends on a meta-parameter $\theta$, biasing gradually the forests probabilities from uniform towards low-cost forests. A sum-over-paths statistical physics framework is used in order to derive the form of the index in terms of the immediate costs defined on the arcs.
It can be computed efficiently by inverting a $n \times n$ matrix, where $n$ is the number of nodes, leading to an overall time complexity of $\mathcal{O}(n^3)$.

The application of the SoF density index to the task of searching dense areas on graphs shows that it performs well, being able to recover all the high density regions -- corresponding to the center of clusters -- on different graphs. Moreover, the correlation results between the SoF density index and the true density (when available) are often close to one. The SoF density index also gives more stable results than the strength regarding the way a graph is constructed. 

In the future, this index could be used together with a density-based clustering method, for instance a mode seeking algorithm on graphs (like in \cite{Cho2012}), for clustering tasks. We will also investigate the application of the proposed technique on large graphs, as in \cite{Mantrach-2011}.  Indeed, the Sum-over-Forests measure only depends on the diagonal of the inverse matrix $\mathbf{Z}$ (this can be easily deduced from Equation (\ref{Eq_Computation_passage03}). Moreover, the matrix $(\mathbf{I} + \mathbf{L}(\mathbf{W}))^{-1}$ is diagonally dominant). In this case, scalable methods can be used for computing the diagonal of $\mathbf{Z}$ (see, e.g., \cite{Duff-1986, Erisman-1975, Tang-2012}).

\ifCLASSOPTIONcompsoc
  \section*{Acknowledgments}
\else
  \section*{Acknowledgment}
\fi
\footnotesize
Part of this work has been funded by projects with the
\textquotedblleft R\'{e}gion wallonne\textquotedblright.
We thank this institution for giving us the opportunity to conduct both
fundamental and applied research.

\ifCLASSOPTIONcaptionsoff
  \newpage
\fi



%

\bibliographystyle{IEEEtran}
\bibliography{Biblio}

\end{document}